\documentclass[letterpaper]{article} 
\usepackage{aaai2026}  
\usepackage{times}  
\usepackage{helvet}  
\usepackage{courier}  
\usepackage[hyphens]{url}  
\usepackage{graphicx} 
\usepackage{natbib}  
\usepackage{caption} 
\usepackage{algorithm}
\usepackage{algorithmic}
\usepackage{amsmath}    
\usepackage{amssymb}    
\usepackage{bm}         
\usepackage{mathtools}     
\usepackage{mathrsfs}      
\usepackage{newfloat}
\usepackage{listings}

\usepackage{subcaption} 

\usepackage{amsfonts}
\usepackage{multirow}
\usepackage{booktabs}
\usepackage{cleveref}
\usepackage{pifont}   
\usepackage{xcolor}
\usepackage{cite}
\newcommand{\cmark}{\textcolor{green!60!black}{\ding{51}}} 
\newcommand{\xmark}{\textcolor{red}{\ding{55}}}            
\newcommand{\fmark}{\textcolor{orange}{\ding{108}}}        

\usepackage{newfloat}
\usepackage{listings}
\DeclareCaptionStyle{ruled}{labelfont=normalfont,labelsep=colon,strut=off} 
\floatstyle{ruled}
\newfloat{listing}{tb}{lst}{}
\floatname{listing}{Listing}
 \nocopyright 

\title{CleanUpBench: Embodied Sweeping and Grasping Benchmark}

\author {
    Wenbo Li\textsuperscript{\rm 1},
    Guanting Chen\textsuperscript{\rm 1},
    Tao Zhao\textsuperscript{\rm 1},
    Jiyao Wang\textsuperscript{\rm 1},\\
    Tianxin Hu\textsuperscript{\rm 2},
    Yuwen Liao\textsuperscript{\rm 2},
    Weixiang Guo\textsuperscript{\rm 2},
    Shenghai Yuan\textsuperscript{\rm 2 *}
}
\affiliations {
    \textsuperscript{\rm 1}Sichuan University, China, \quad 
    \textsuperscript{\rm 2}Nanyang Technological University, Singapore\\
    \texttt{
    liwenbo1@stu.scu.edu.cn,\quad 
    shyuan@ntu.edu.sg
    }
}

\usepackage{bibentry}

\begin{document}

\maketitle

\begin{abstract}
Embodied AI benchmarks have advanced navigation, manipulation, and reasoning, but most target complex humanoid agents or large-scale simulations that are far from real-world deployment.  In contrast, mobile cleaning robots with dual mode capabilities, such as sweeping and grasping, are rapidly emerging as realistic and commercially viable platforms. However, no benchmark currently exists that systematically evaluates these agents in structured, multi-target cleaning tasks, revealing a critical gap between academic research and real-world applications. We introduce CleanUpBench, a reproducible and extensible benchmark for evaluating embodied agents in realistic indoor cleaning scenarios. Built on NVIDIA Isaac Sim, CleanUpBench simulates a mobile service robot equipped with a sweeping mechanism and a six-degree-of-freedom robotic arm, enabling interaction with heterogeneous objects. The benchmark includes manually designed environments and one procedurally generated layout to assess generalization, along with a comprehensive evaluation suite covering task completion, spatial efficiency, motion quality, and control performance. To support comparative studies, we provide baseline agents based on heuristic strategies and map-based planning. CleanUpBench bridges the gap between low-level skill evaluation and full-scene testing, offering a scalable testbed for grounded, embodied intelligence in everyday settings. All code and benchmarks will be released as open source upon acceptance.

\end{abstract}

%
\begin{links}
\end{links}

\section{Introduction}
\begin{figure}[t]
\centering
\includegraphics[width=\linewidth]{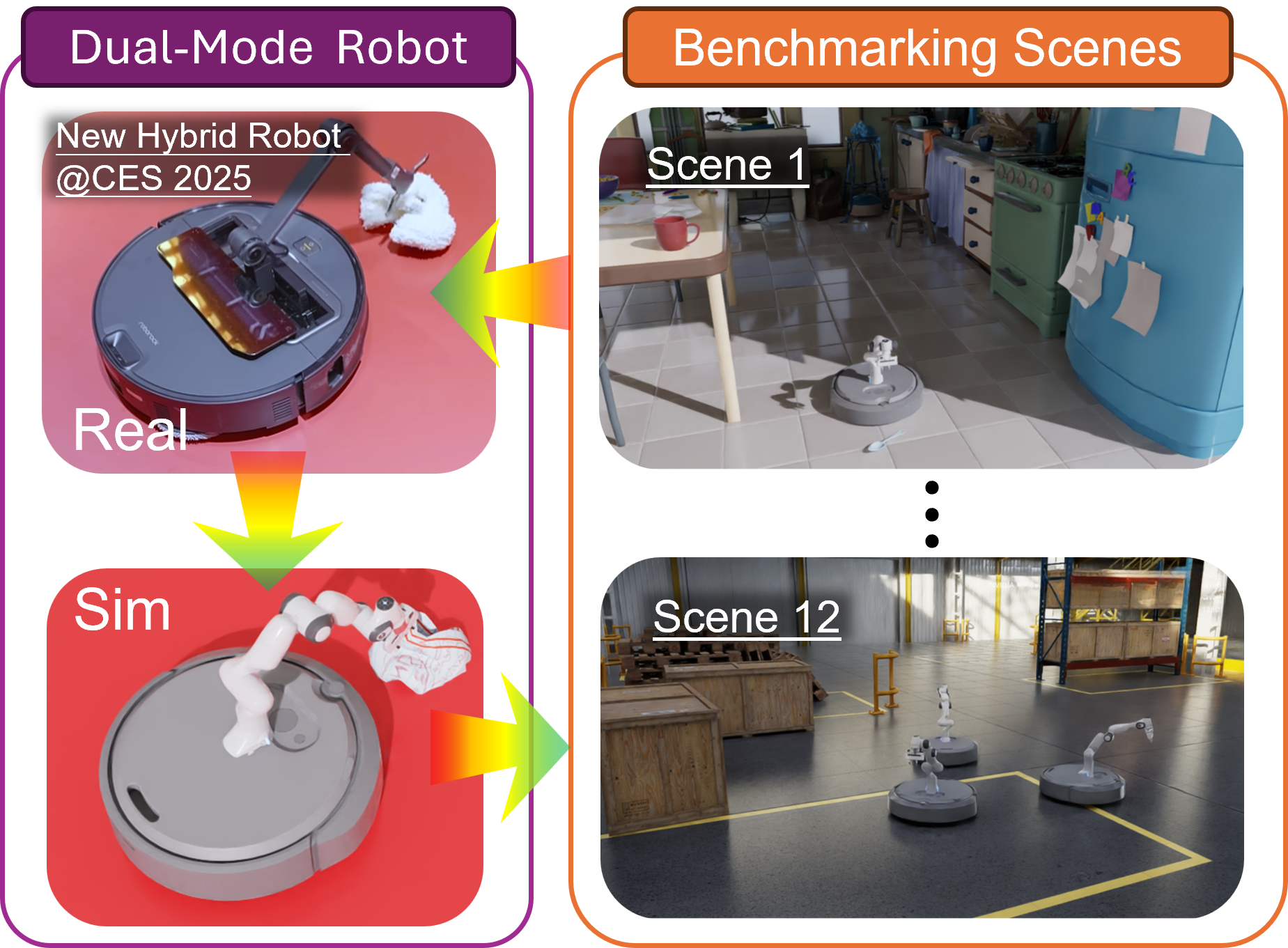}  
\caption{CleanUpBench motivation and system overview: (Top) Dual-mode robot platform bridging real hardware demonstrated at CES 2025 to simulation environment. (Down) Benchmarking scenes including diverse indoor layouts for systematic evaluation of cleaning agents.}
\label{fig:motivations}
\end{figure}

In recent years, \textbf{Embodied AI} has emerged as a key research frontier in artificial intelligence, with growing applications in household services~\cite{shridhar2020alfred, yenamandra2023homerobot}, warehouse logistics~\cite{jaafar2024lambda}, and assistive healthcare~\cite{padmakumar2021teach}. Among these, mobile robotic systems designed for \textbf{multi-target interactive tasks}, particularly cleaning robots equipped with intelligent planning and execution capabilities, have drawn increasing attention~\cite{jiang2025brs}. For example, service robots in real homes are expected to perform integrated behaviors such as autonomous navigation, object recognition, and task-specific manipulation through \textbf{dual-mode interaction} (e.g., picking up clutter or sweeping debris), as shown in Fig. \ref{fig:motivations}. However, compared to static perception or navigation tasks, these \textbf{mobile dual-mode interaction} scenarios remain underexplored in terms of unified research frameworks and reproducible evaluation standards~\cite{savva2019habitat, mees2023calvin}, thereby limiting the deployment of \textbf{embodied agents} in real world settings.

Current embodied AI benchmarks tend to exhibit a \textbf{polarized landscape}. On one side, existing platforms focus on isolated skills such as navigation (e.g., ObjectNav~\cite{chaplot2020objectnav}, PointNav~\cite{anderson2018vision}), object manipulation (e.g., RLBench~\cite{james2020rlbench}), or spatial reconstruction (e.g., BEHAVIOR~\cite{jiang2025brs}), offering well defined tasks but limited task synergy. On the other side, simulation frameworks such as RobotCasa emulate highly complex multi step household routines involving perception, language, and control~\cite{chang2025partnr, padmakumar2021teach}. While holistic, these systems often involve high development costs and steep learning curves~\cite{deitke2020robothor}, making them impractical for evaluating specific capabilities. For \textbf{embodied cleaning tasks}, which are goal driven and moderately complex, existing platforms fall short: they either lack unified modeling and evaluation of dual-mode interaction (e.g., sweeping and grasping), or fail to provide controllable environments for verifying cleaning strategies and target distribution across multiple robot configurations~\cite{mees2023calvin, jaafar2024lambda}. This reveals a critical need for a \textbf{task-centric, reproducible, and extensible evaluation platform} that balances between minimalistic skill assessment and complete household simulations.

Designing a benchmark for cleaning robots in realistic indoor environments introduces several challenges. First, the environment must exhibit \textbf{structural diversity and occlusion}, reflecting real world difficulties in path planning and object perception~\cite{xia2020interactive}. Second, the task design should integrate \textbf{dual-mode interaction objectives}, requiring the agent to perform \textbf{dynamic task scheduling and behavioral switching}~\cite{zhang2023mapthor, chang2025partnr}. Third, the evaluation metrics should comprehensively assess \textbf{task success, path efficiency, and interaction quality}~\cite{jiang2025brs, mees2023calvin}, in order to prevent overfitting to any single aspect. Finally, the platform should be \textbf{reproducible, lightweight, and broadly compatible} with various control policies, including both heuristic planners and learned agents, to support comparative studies and promote general purpose methods~\cite{savva2019habitat, james2020rlbench}.

To address these gaps, we introduce \textbf{CleanUpBench}, a new benchmark designed for evaluating \textbf{embodied cleaning agents}. CleanUpBench simulates the full workflow of cleaning tasks in domestic settings with time constraints, including autonomous exploration, sweeping loose debris, and grasping large objects through dual-mode interaction. Built on the NVIDIA Isaac Sim engine, our platform provides \textbf{highly controllable and diverse scene configurations} with 20 scenes spanning 5 distinct categories, supporting both single-robot operations and up to 3-robot collaborative scenarios. We propose a \textbf{comprehensive evaluation suite} that measures task completion, navigation redundancy, and interaction efficiency. Additionally, we provide several \textbf{baseline agents}, ranging from greedy sweeping strategies to map based A* planners and low-level action controllers, as standard comparison tools. We envision \textbf{CleanUpBench} as a \textbf{realistic, scalable, and reproducible testbed} that supports the development and evaluation of intelligent service robots in human-centric environments.

In summary, our key contributions are as follows:

\begin{itemize}
    \item We present \textbf{CleanUpBench}, a novel embodied AI benchmark that targets realistic and structured household cleaning tasks through dual-mode interaction, addressing the gap between low-level skill testing and high-level system integration.

    \item We develop a physics-accurate simulation platform based on \textbf{NVIDIA Isaac Sim}, featuring mobile cleaning robots with both sweeping modules and robotic grippers, supporting single-robot or 3-robot collaborative operations.

    \item We propose a \textbf{comprehensive evaluation protocol} with multi-level metrics that assess spatial coverage, task success through dual-mode interaction, motion quality, safety performance, and computational efficiency.

    \item We construct a diverse set of \textbf{20 test environments across 5 distinct categories}, including manually designed indoor layouts and procedurally generated scenes, to support both performance benchmarking and generalization evaluation.

    \item We implement several \textbf{baseline agents}, including heuristic policies and map-based planners, demonstrating the benchmark's modularity and serving as reference points for future learning-based and hybrid approaches.
\end{itemize}


\begin{table*}[t]
\centering

\renewcommand{\arraystretch}{1.1}
\setlength{\tabcolsep}{1.5mm}
\begin{tabular}{l|ccccccc}
\hline
\textbf{Benchmark} & \textbf{Dual-Mode} & \textbf{Physics Sim} & \textbf{Layout Control} & \textbf{Scene Gen} & \textbf{Rich Metrics} & \textbf{Cluttered Env} & \textbf{Multi-Agent} \\
\hline
AI2-THOR      & \fmark & \fmark & \cmark & \cmark & \fmark & \cmark & \xmark \\
Habitat       & \xmark & \fmark & \cmark & \cmark & \fmark & \cmark & \fmark \\
RLBench       & \cmark & \cmark & \xmark & \xmark & \fmark & \fmark & \xmark \\
BEHAVIOR      & \cmark & \fmark & \xmark & \fmark & \fmark & \cmark & \xmark \\
ALFRED        & \xmark & \fmark & \fmark & \xmark & \fmark & \cmark & \xmark \\
CALVIN        & \cmark & \cmark & \cmark & \fmark & \fmark & \fmark & \xmark \\
CleanUpBench  & \cmark & \cmark & \cmark & \cmark & \cmark & \cmark & \cmark  \\
\hline
\end{tabular}
\caption{Benchmark capability comparison across 7 key embodied intelligence dimensions. \cmark\ Supported\quad
\xmark\ Not supported\quad
\fmark\ Partially supported.
\textbf{Dual-Mode}: Supports both sweeping and grasping interaction. \quad
\textbf{Physics Sim}: Uses accurate physics (e.g., Isaac, Mujoco). 
\textbf{Layout Control}: Allows precise object/obstacle placement. \quad
\textbf{Scene Gen}: Supports procedural/randomized environments. 
\textbf{Rich Metrics}: Includes motion, success, timing, and smoothness evaluation. \quad
\textbf{Cluttered Env}: Contains dense occlusions or obstacles. 
\textbf{Multi-Agent}: Supports multi-robot or cooperative tasks.}
\raggedright
\footnotesize
\label{table_comparsion}
\end{table*}

\section{Related Works}
\textbf{Embodied AI Benchmarks.} Interactive environments such as AI2-THOR~\cite{kolve2017ai2thor}, Habitat~\cite{savva2019habitat}, and iGibson~\cite{xia2020interactive} have advanced indoor navigation and object interaction research. Task-specific benchmarks like ObjectNav~\cite{chaplot2020objectnav}, RLBench~\cite{james2020rlbench}, and BEHAVIOR~\cite{jiang2025brs} focus on isolated skills but lack integrated dual-mode interaction evaluation.

\textbf{Recent Comprehensive Benchmarks.} EMMOE~\cite{li2025emmoe}, EmbodiedBench~\cite{yang2025embodiedbench}, and EmbodiedEval~\cite{cheng2025embodiedeval} provide comprehensive evaluation platforms but primarily target single-interaction modalities without coordinated dual-mode behaviors.

\textbf{Household and Service Robotics.} TEACh~\cite{padmakumar2021teach}, ALFRED~\cite{shridhar2020alfred}, and HomeRobot~\cite{yenamandra2023homerobot} incorporate natural language and household tasks but focus mainly on single-modality interactions rather than integrated sweeping-grasping coordination.

Detailed benchmark comparisons are provided in Supplementary Material Section B. CleanUpBench addresses the gap by explicitly supporting dual-mode interaction evaluation across diverse scenarios with comprehensive metrics.

\section{Benchmark Design}

\textbf{CleanUpBench} is a high-fidelity simulation benchmark designed to evaluate embodied cleaning agents operating in realistic and cluttered home-like environments. Built upon \textbf{NVIDIA Isaac Sim}, the platform leverages advanced photorealistic rendering, sensor simulation (e.g., RGB-D, segmentation, LiDAR), and accurate rigid-body physics. These features provide an ideal testbed for embodied AI research by bridging the gap between synthetic simulation and real-world deployment.

The central robotic agent is a wheeled service robot equipped with two modes of physical interaction: a front-mounted \textbf{sweeping module} (e.g., vacuum-like roller or brush) and a \textbf{6-DOF robotic manipulator with a parallel gripper}. This dual-mode interaction configuration enables the agent to handle a wide range of objects in indoor cluttered scenes, including \textit{sweepable} debris (e.g., paper scraps, dirt, dust piles) and \textit{graspable} items (e.g., bottles, toys, remote controls). The platform supports both single-robot operations and collaborative scenarios with up to 3 robots working coordinately.

The simulation includes \textbf{20 manually designed scenes} spanning 5 distinct complexity categories (4 per category), plus \textbf{procedural generation} capabilities to systematically evaluate different aspects of embodied cleaning intelligence. Each scene targets specific behavioral competencies:
The simulation includes 20 scenes across 5 distinct categories designed to test different behavioral competencies. We show a rough comparison with other benchmarks as shown in Tab. \ref{table_comparsion}.   Detailed scene specifications are provided in Supplementary Material Section A.1.

\subsection{Procedural Scene Generation}
Our procedural generation system creates diverse environments by varying room layouts, obstacle density, and target distribution patterns, enabling generalization evaluation beyond the 20 fixed scenes. Details are provided in Supplementary Material Section A.4.
    
The benchmark supports multiple control paradigms:
\begin{itemize}
    \item \textbf{Optimization-based methods}: e.g., A* or D* planning over occupancy grids.
    \item \textbf{Heuristic or rule-based controllers}: e.g., finite state machines for strategy switching.
    \item \textbf{Learning-based policies}: e.g., RL agents trained with PPO, SAC, or imitation learning.
\end{itemize}
All agents can interact with the simulation via a standardized control interface that exposes proprioception, vision, and actuator APIs for both single-robot and multi-robot configurations. This promotes fair comparisons between method families under identical conditions.

\section*{Evaluation Protocol}

Each episode starts by initializing a cleaning robot at a randomly selected spawn location in a given scene. A fixed set of debris and objects are spawned at randomized positions. The agent must complete the cleaning task within a pre-defined time budget \( T_{\text{max}} \) while maximizing task completion and minimizing redundant or inefficient behavior.

Agents operate in two main settings:
\begin{itemize}
    \item \textbf{Seen scenarios}: training and evaluation scenes drawn from the same fixed scene pool.
    \item \textbf{Unseen scenario}: a procedurally generated scene withheld from training to test spatial and generalization
\end{itemize}

Agents receive either low-level observations (e.g., depth image, joint states) or high-level perception (e.g., segmentation masks, affordance maps), depending on the task mode. Episodes are terminated upon timeout, full task completion, or agent failure.

\begin{figure*}[t]
  \centering
  \includegraphics[width=0.95\linewidth]{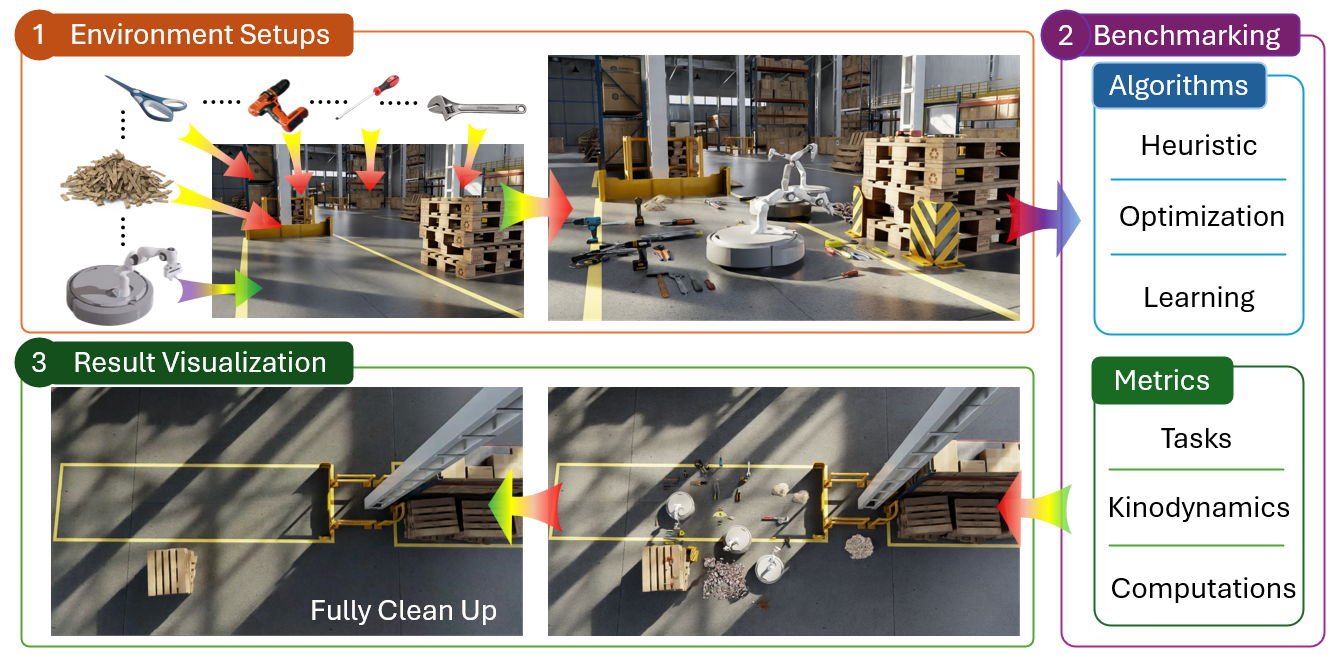}
  \caption{System overview of CleanUpBench: The agent performs dual-mode interaction—sweeping and grasping—under spatially diverse and procedurally generated indoor scenes.}
  \label{fig:flowchart}
\end{figure*}

\section*{Metrics}

\subsection*{Evaluation Framework}

We establish a comprehensive evaluation framework for embodied cleaning agents as shown in Fig. \ref{fig:flowchart}. Complete notation and symbol definitions are provided in Supplementary Material Section H. Let $\mathcal{E} = (\mathcal{S}, \mathcal{A}, \mathcal{T})$ represent the cleaning environment, where $\mathcal{S}$ is the state space, $\mathcal{A}$ is the action space, and $\mathcal{T}: \mathcal{S} \times \mathcal{A} \rightarrow \mathcal{S}$ is the transition function. An episode consists of a temporal sequence $\{s_\tau, a_\tau\}_{\tau=1}^{\tau_{\max}}$, where $s_\tau \in \mathcal{S}$ represents the environment state and $a_\tau \in \mathcal{A}$ represents the agent's action at time step $\tau$.

The robot's physical representation is modeled as a rigid body with footprint $\mathcal{F} \subset \mathbb{R}^2$ in the horizontal plane. At each time step $\tau$, the robot occupies position $x_\tau = (x_\tau^{(1)}, x_\tau^{(2)}) \in \mathbb{R}^2$ with orientation $\theta_\tau \in SO(2)$. The robot's configuration space trajectory is denoted as $\mathcal{C} = \{(x_\tau, \theta_\tau)\}_{\tau=1}^{\tau_{\max}}$.

We define a comprehensive metric suite $\mathbb{M} = \{\text{CR}, \text{TCR}, \text{ME}, \text{SR}, \text{Collision}, \text{CT}, \text{FT}, \text{Vel}_{\text{avg}}, \text{Acc}_{\text{avg}}, \text{Jerk}_{\text{avg}}\}$ to quantitatively evaluate performance across task success, planning efficiency, motion smoothness, and stability.

\subsection*{Spatial Coverage Metrics}

\subsubsection*{Coverage Ratio (CR)}

The \textbf{Coverage Ratio} quantifies the proportion of navigable space explored by the robot during task execution:
\[
\text{CR} = \frac{A_{\text{covered}}}{A_{\text{total}}}
\]

\noindent where $A_{\text{total}}$ represents the total navigable floor area in the environment, defined as:
\[
A_{\text{total}} = \int_{\mathcal{W}} \mathbb{I}[\text{navigable}(p)] \, dp
\]

\noindent Here, $\mathcal{W} \subset \mathbb{R}^2$ denotes the workspace boundary and $\mathbb{I}[\text{navigable}(p)]$ is an indicator function that equals 1 if point $p$ is collision-free and accessible.

$A_{\text{covered}}$ represents the cumulative area swept by the robot's chassis during motion:
\[
A_{\text{covered}} = \text{Area}\left(\bigcup_{\tau=1}^{\tau_{\max}} \mathcal{F}_\tau\right)
\]

\noindent where $\mathcal{F}_\tau = \{p \in \mathbb{R}^2 : p \in \mathcal{F} \oplus (x_\tau, \theta_\tau)\}$ is the robot footprint at configuration $(x_\tau, \theta_\tau)$ and $\oplus$ denotes the Minkowski sum operation for geometric transformation.

A higher CR indicates better spatial exploration effectiveness, while lower values suggest incomplete or inefficient coverage.

\subsubsection*{Sweep Redundancy (SR)}

The \textbf{Sweep Redundancy} measures spatial efficiency by quantifying how much area was unnecessarily re-swept:
\[
\text{SR} = \frac{\sum_{g \in \mathbb{G}} \mathbb{I}[\nu(g) > 1]}{\sum_{g \in \mathbb{G}} \mathbb{I}[\nu(g) \geq 1]}, \quad \nu(g) = \sum_{\tau=1}^{\tau_{\max}} \mu_\tau(g)
\]

\noindent Here, $\mathbb{G}$ represents a discrete grid decomposition of the environment floor with resolution $\delta$:
\[
\mathbb{G} = \{g_{i,j} = (i\delta, j\delta) : i,j \in \mathbb{Z}, g_{i,j} \in \mathcal{W}\}
\]

\noindent $\mu_\tau(g) \in \{0,1\}$ is a binary indicator function:
\[
\mu_\tau(g) = \mathbb{I}[g \cap \mathcal{F}_\tau \neq \emptyset]
\]

\noindent which equals 1 if grid cell $g$ intersects with the robot footprint at time $\tau$. The visit count $\nu(g)$ accumulates how many times cell $g$ was swept throughout the episode.

Lower SR values are desirable, indicating minimal spatial overlap and higher efficiency in area coverage.

\subsection*{Task Performance Metrics}

\subsubsection*{Task Completion Ratio (TCR)}

The \textbf{Task Completion Ratio} evaluates the agent's effectiveness in dual-mode interaction with cleanable objects through a decomposed scoring mechanism:
\[
\text{TCR} = \alpha \cdot \text{TCR}_{\text{sweep}} + \beta \cdot \text{TCR}_{\text{grasp}}
\]
\noindent where $\alpha + \beta = 1$ and typically $\alpha = \beta = 0.5$ for balanced evaluation. The individual components are defined as:
\[
\text{TCR}_{\text{sweep}} = \frac{N_{\text{S-success}}}{N_{\text{S-total}}}, \quad \text{TCR}_{\text{grasp}} = \frac{N_{\text{G-success}}}{N_{\text{G-total}}}
\]

\noindent Let $\mathcal{O}_S$ and $\mathcal{O}_G$ represent the sets of sweepable and graspable objects in the environment, respectively. Then:
\begin{itemize}
    \item $N_{\text{S-total}} = |\mathcal{O}_S|$ and $N_{\text{G-total}} = |\mathcal{O}_G|$ are the total counts
    \item $N_{\text{S-success}} = |\{o \in \mathcal{O}_S : \text{swept}(o) = \text{true}\}|$ and $N_{\text{G-success}} = |\{o \in \mathcal{O}_G : \text{grasped}(o) = \text{true}\}|$ are the successfully completed counts
\end{itemize}

This decomposed scoring allows fair evaluation of algorithms with different capabilities: agents supporting dual-mode interaction receive full TCR scores, while single-mode algorithms are evaluated only on their applicable tasks (TCR$_{\text{sweep}}$ for coverage-only methods, TCR$_{\text{grasp}}$ for manipulation-only methods).

\subsubsection*{Motion Efficiency (ME)}

The \textbf{Motion Efficiency} measures path efficiency per completed interaction:
\[
\text{ME} = \frac{L_{\text{total}}}{N_{\text{S-success}} + N_{\text{G-success}}}
\]

\noindent where $L_{\text{total}}$ is the robot's total trajectory path length in Euclidean space:
\[
L_{\text{total}} = \sum_{\tau=1}^{\tau_{\max}-1} \|x_{\tau+1} - x_\tau\|_2
\]

\noindent This metric reflects the average distance traveled per successful object interaction. Lower ME values are preferred, indicating efficient motion planning that minimizes unnecessary travel.

\subsection*{Safety and Robustness Metrics}

\subsubsection*{Collision Count}

The \textbf{Collision} metric accumulates safety violations throughout the episode:
\[
\text{Collision} = \sum_{\tau=1}^{\tau_{\max}} \mathbb{I}[\chi_\tau = \top]
\]

\noindent where $\chi_\tau$ is a binary collision indicator:
\[
\chi_\tau = \mathbb{I}[\mathcal{F}_\tau \cap \mathcal{O}_{\text{static}} \neq \emptyset]
\]

\noindent Here, $\mathcal{O}_{\text{static}}$ represents the union of all static obstacles (walls, furniture, fixtures) in the environment. Lower collision counts indicate safer and more robust navigation.

\subsection*{Computational Performance Metrics}

\subsubsection*{Computation Time (CT)}

The \textbf{Computation Time} evaluates the average computational time per control decision:
\[
\text{CT} = \frac{1}{\tau_{\max}} \sum_{\tau=1}^{\tau_{\max}} t_{\text{comp}}(\tau)
\]

\noindent where $t_{\text{comp}}(\tau)$ is the wall-clock time required to compute the control action $a_\tau$ given state $s_\tau$. Lower CT values indicate better real-time feasibility for embedded deployment.

\subsubsection*{Finish Time (FT)}

The \textbf{Finish Time} measures total episode execution duration:
\[
\text{FT} = \tau_{\text{final}} - \tau_{\text{init}}
\]

\noindent where $\tau_{\text{init}}$ and $\tau_{\text{final}}$ represent the episode start and end time steps, respectively. This metric should be balanced against task quality measures, as faster completion may compromise thoroughness.

\subsection*{Motion Quality Metrics}

The motion smoothness is characterized by the statistical properties of the robot's kinematic derivatives. Let $x_\tau \in \mathbb{R}^2$ denote the robot's position at time $\tau$ with discrete time interval $\Delta t$.

\subsubsection*{Average Velocity}
\[
\text{Vel}_{\text{avg}} = \frac{1}{\tau_{\max}} \sum_{\tau=1}^{\tau_{\max}} \left\| \frac{x_{\tau+1} - x_\tau}{\Delta t} \right\|_2
\]

\subsubsection*{Average Acceleration}
\[
\text{Acc}_{\text{avg}} = \frac{1}{\tau_{\max}} \sum_{\tau=1}^{\tau_{\max}} \left\| \frac{(x_{\tau+1} - x_\tau) - (x_\tau - x_{\tau-1})}{(\Delta t)^2} \right\|_2
\]

\subsubsection*{Average Jerk}
\[
\text{Jerk}_{\text{avg}} = \frac{1}{\tau_{\max}} \sum_{\tau=1}^{\tau_{\max}} \left\| \frac{a_{\tau+1} - a_\tau}{\Delta t} \right\|_2
\]

\noindent where $a_\tau$ represents the discrete acceleration at time $\tau$.

Lower values across all kinematic metrics indicate smoother and more stable motion profiles, which are favorable for mechanical wear reduction, energy efficiency, and passenger comfort in mobile robotics applications.

\begin{table*}[t]
\centering
\caption{Baseline method performance on CleanUpBench. Results averaged across 5 runs. Methods are categorized by capability: Sweep-only, Grasp-only, or Dual-mode. Bold indicates best performance per column.}
\label{tab:example_methods}
\renewcommand{\arraystretch}{1.1}

\setlength{\tabcolsep}{1.5mm} 
\begin{tabular}{l|c|c|c|c|cccccccccc}
\hline
\hline
\textbf{Method} & \textbf{Year} & \textbf{Type} & \textbf{Mode} & \textbf{Robots} & \textbf{TCR} & \textbf{TCR$_S$} & \textbf{TCR$_G$} & \textbf{ME} & \textbf{SR} & \textbf{CR} & \textbf{FT} & \textbf{CT} & \(\textbf{Vel}_{\text{avg}}\) & \textbf{Col} \\
\hline
m-explore  & 2016 & Heuristic & Sweep & Single & 0.03 & 0.05 & 0.00 & 0.04 & 0.03 & 0.08 & \textbf{101.99}& \textbf{0.01}& 0.09 & \textbf{0.01}\\
Manhattan  & 2021 & Heuristic & Sweep & Single & 0.12 & 0.23 & 0.00 & 0.08 & \textbf{0.01} & 0.34 & 250.20 & 0.08 & 0.16 & 8.00 \\
Chebyshev  & 2021 & Heuristic & Sweep & Single & 0.15 & 0.30 & 0.00 & 0.10 & \textbf{0.01} & 0.42 & 249.66 & 0.08 & 0.19 & 6.00 \\
Vertical  & 2021 & Heuristic & Sweep & Single & 0.12 & 0.24 & 0.00 & 0.08 & \textbf{0.01} & 0.23 & 250.17 & 0.08 & 0.19 & 7.00 \\
Horizontal  & 2021 & Heuristic & Sweep & Single & 0.15 & 0.29 & 0.00 & 0.07 & \textbf{0.01} & 0.36 & 249.83 & 0.08 & 0.15 & 6.00 \\
CA-DRL  & 2023 & RL & Sweep & Single & 0.00 & 0.00 & 0.00 & 0.05 & 0.63 & 0.54 & 249.30 & 0.34 & \textbf{1.12} & 81.00 \\
IR2  & 2024 & RL & Sweep & Multi & 0.17 & 0.33 & 0.00 & 0.01& 1.00 & 0.37 & 249.00& 0.34 & 0.33 & 2.00 \\
\hline
BHyRL  & 2022 & RL & Grasp & Single & 0.20 & 0.00 & 0.40 & 0.07 & 0.99 & 0.37 & 313.14 & 0.25 & 0.36 & 1.00 \\
REMANI  & 2024 & Planning & Grasp & Single & 0.30 & 0.00 & 0.60 & \textbf{0.23}& 1.57 & 0.57 & 210.01& 0.02 & 0.12 & 2.00 \\
\hline
PRIMAL2  & 2021 & RL & Dual & Multi & \textbf{0.60} & \textbf{0.50} & \textbf{0.70} & 0.21 & \textbf{0.01}& \textbf{0.69} & 287.73 & 0.04& 0.20 & 0.04\\
\hline
\end{tabular}
\vspace{2mm}
\begin{flushleft}
\footnotesize
\textbf{TCR}: Overall Task Completion Rate. \textbf{TCR$_S$}: Sweep Task Completion Rate. \textbf{TCR$_G$}: Grasp Task Completion Rate. \textbf{ME}: Motion Efficiency (m/target). \textbf{SR}: Sweep Redundancy. \textbf{CR}: Coverage Rate. \textbf{FT}: Task Completion Time (s). \textbf{CT}: Computation Time (s). \(\textbf{Vel}_{\text{avg}}\): Average Velocity (m/s). \textbf{Col}: Total Collision Count.
\end{flushleft}
\end{table*}

\begin{figure*}[t]
\centering
\includegraphics[width=\textwidth]{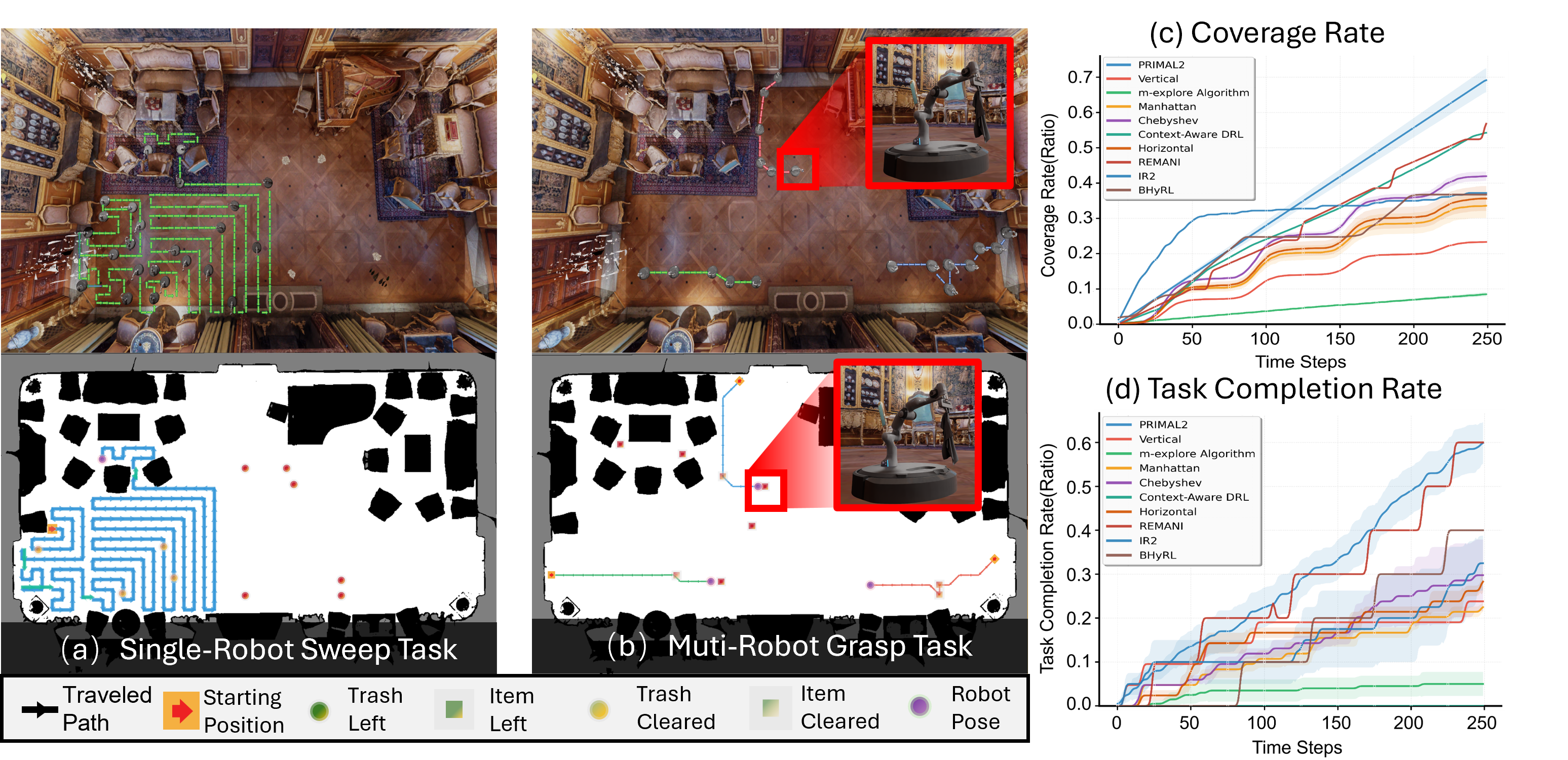}
\caption{Single-robot sweep task (a) vs. multi-robot grasp task (b) with coordinated path planning. (c) and (d) are selected cumulative performance evaluations based on our proposed metrics systems.  }
\label{fig:map}
\end{figure*}

\section*{Experimental Setup}
We evaluate ten baseline methods spanning classical planning, heuristic strategies, and learning-based control to demonstrate the diversity and extensibility of \textbf{CleanUpBench}. These baselines are chosen to reflect a broad range of policy designs in embodied intelligence, from fully reactive schemes to globally optimized planners supporting different dual-mode interaction capabilities.

\begin{itemize}
    \item \textbf{Manhattan \cite{tan2021comprehensive}} uses the Manhattan distance heuristic for coverage planning, allowing movement in four cardinal directions. It offers deterministic behavior but lacks diagonal efficiency. We include it as a simple, established baseline for single-robot sweeping tasks in structured environments. (\textit{Sweep-only, Single-robot})
    
    \item \textbf{Chebyshev \cite{tan2021comprehensive}} uses the Chebyshev distance heuristic to enable 8-directional movement, improving coverage efficiency over Manhattan while remaining computationally simple. It serves as a baseline for single-robot sweeping tasks in grid-based environments. (\textit{Sweep-only, Single-robot})

    \item \textbf{Vertical \cite{tan2021comprehensive}} implements a systematic vertical sweeping pattern for complete area coverage. This method ensures predictable coverage behavior and minimal overlap through deterministic path planning. (\textit{Sweep-only, Single-robot})
    
    \item \textbf{Horizontal \cite{tan2021comprehensive}} performs systematic horizontal sweeping for comprehensive area coverage, providing consistent coverage with minimal redundancy through systematic movement patterns. (\textit{Sweep-only, Single-robot})
    
    \item \textbf{m-explore \cite{horner2016mapmerging}} is a frontier-based exploration algorithm originally developed for multi-robot exploration tasks. It identifies and navigates to unexplored boundaries using frontier detection and selection strategies. We adapt it as a baseline for its proven exploration capabilities. (\textit{Sweep-only, Single-robot})
    
    \item \textbf{PRIMAL2 \cite{damani2021primal}} is a decentralized reinforcement learning approach for lifelong multi-agent path finding in dense environments. It learns reactive policies under partial observability and scales to multiple agents. We select it as a baseline for its strong coordination ability and scalability in dual-mode interaction scenarios. (\textit{Dual-mode, Multi-robot})
    
    \item \textbf{IR2 \cite{tan2024ir2}} leverages attention-based neural networks for multi-robot exploration under sparse intermittent connectivity. It enables implicit coordination decisions by reasoning about long-term trade-offs between solo exploration and information sharing. (\textit{Sweep-only, Multi-robot})
    
    \item \textbf{CA-DRL \cite{liang2023context}} employs a context-aware policy network for mapless navigation in unknown environments. It forms contextual beliefs over the entire known area to reason about long-term efficiency and plan short-term movements. (\textit{Sweep-only, Single-robot})
    
    \item \textbf{BHyRL \cite{jauhri2022robot}} uses Boosted Hybrid Reinforcement Learning for mobile manipulation with reachability behavior priors. It combines discrete base placement decisions with continuous arm control through hybrid action spaces. (\textit{Grasp-only, Single-robot})
    
    \item \textbf{REMANI-Planner \cite{wu2024remani}} implements real-time whole-body motion planning for mobile manipulators using environment-adaptive search and spatial-temporal optimization. It generates collision-free trajectories for manipulation tasks. (\textit{Grasp-only, Single-robot})
\end{itemize}

These baselines vary across three key dimensions: \textit{interaction capability} (sweep-only, grasp-only, or dual-mode), \textit{robot configuration} (single-robot or multi-robot), and \textit{algorithmic approach} (heuristic, planning-based, or learning-based). The decomposed TCR evaluation allows fair comparison by assessing each method only on its applicable interaction modes.

Each method is evaluated across all 20 scenes spanning 5 distinct categories, with 5 independent runs per scene to ensure statistical reliability. All runs use identical time limits (300s), robot configurations, and sensor modalities. Metrics follow the comprehensive framework defined in Section \textbf{Metrics}, with particular emphasis on the dual-mode interaction capabilities of each approach.

\subsection{Result and Analysis}

We evaluate ten baseline methods on CleanUpBench across 20 diverse scenes spanning 5 distinct categories. Table~\ref{tab:example_methods} shows results spanning heuristic and learning-based approaches with different dual-mode interaction capabilities.

\noindent\textbf{Overall Performance.} PRIMAL2 achieves the best overall performance with TCR=0.60, demonstrating superior dual-mode coordination capabilities (TCR$_S$=0.50, TCR$_G$=0.70). Traditional sweep-only methods like Chebyshev show moderate single-mode performance (TCR$_S$=0.30) but zero grasping capability, while grasp-only methods like REMANI achieve TCR$_G$=0.60 but cannot perform sweeping tasks. This reveals the fundamental advantage of dual-mode interaction systems over single-mode approaches.

\noindent\textbf{Dual-Mode Coordination Analysis.}The decomposed TCR evaluation reveals critical insights about dual-mode interaction strategies. PRIMAL2's success stems from its multi-agent coordination principles effectively managing mode transitions and spatial task allocation, achieving balanced performance across both interaction modes. In contrast, single-mode algorithms face inherent limitations: sweep-only methods (Manhattan, Chebyshev, Vertical, Horizontal) achieve excellent sweep redundancy (SR=0.01) through systematic coverage but completely fail at object manipulation tasks. grasp-only methods (BHyRL, REMANI) can handle individual objects but show poor spatial coverage and high sweep redundancy ($\text{SR}\geq0.99$), indicating inefficient exploration patterns. 

\noindent\textbf{Multi-Robot vs Single-Robot Performance.} Multi-robot capable methods (IR2, PRIMAL2) demonstrate distinct advantages in task completion time and coverage efficiency. PRIMAL2's multi-robot coordination enables parallel task execution across different interaction modes, while IR2 shows perfect motion efficiency (ME=0.00) through collaborative exploration. However, coordination complexity increases computational overhead (CT=0.34s for multi-robot vs 0.08s for single-robot heuristics).

\noindent\textbf{Heuristic vs Learning Methods.} Classical heuristic methods show consistent but limited performance. All sweep-focused heuristics achieve identical sweep redundancy (SR=0.01) and similar finish times ($\approx$250s), indicating systematic but inflexible behavior patterns. Learning methods exhibit greater performance variance: PRIMAL2 excels through coordination mechanisms, while CA-DRL fails completely (TCR=0.00) with 81 collisions, suggesting that dual-mode interaction requires careful algorithm design for safety and coordination.

\noindent\textbf{Motion Quality and Safety.} Learning methods generally achieve higher velocities but with varying safety profiles. CA-DRL reaches 1.12 m/s but suffers poor control stability and frequent collisions. PRIMAL2 demonstrates superior balance with 0.20 m/s velocity and minimal collisions (0.40), indicating effective dual-mode coordination includes motion safety considerations. Heuristic methods maintain smooth motion profiles but operate at conservative speeds.

\noindent\textbf{Computational Efficiency.} PRIMAL2 achieves the most efficient computation time (0.04s), enabling real-time dual-mode decision making. Traditional methods require moderate computation (0.08s), while complex learning approaches like IR2 and CA-DRL demand significantly more time (0.34s), affecting real-time deployment feasibility for collaborative dual-mode interaction scenarios.

\noindent\textbf{Key Insights.} The dual-mode cleaning task reveals fundamental challenges in embodied AI coordination as shown in Fig. \ref{fig:map}. Methods designed for single interaction modes struggle with task integration and spatial efficiency. Pure coverage approaches work systematically but miss manipulation opportunities. Learning methods show promise for dual-mode coordination but require careful design for safety and stability. Multi-agent coordination principles (as demonstrated by PRIMAL2) appear most effective for managing different interaction modes and spatial task allocation in complex cleaning scenarios.

\section{Limitations and Future Opportunities}
Limitations and Future Opportunities
CleanUpBench offers a high-fidelity benchmark that supports dual-mode interaction, diverse scene layouts, and standardized evaluation. While the current platform already captures many real-world complexities, further opportunities include incorporating dynamic sensory disturbances, richer physical interaction, and large-scale multi-agent coordination. Additional discussions on future extensions and deployment considerations are provided in the Supplementary Material (Sections F and G).

\section{Conclusion}
We introduced \textit{CleanUpBench}, a comprehensive benchmark for evaluating embodied agents in realistic, dual-mode cleaning tasks. By bridging single-skill assessment and complex multi-modal execution, our platform addresses a key gap in embodied AI evaluation.
Through systematic evaluation across eight representative methods, we observe clear trade-offs between structured heuristics and adaptive, coordinated agents. Our metric suite—covering spatial coverage, task success, motion quality, and computational efficiency—enables fine-grained analysis of algorithmic behavior.
With modular design, procedural scene generation, and open-source access, \textit{CleanUpBench} provides a scalable testbed for future research. We expect it to facilitate reproducible studies and accelerate the development of robust, generalizable embodied intelligence for real-world service robotics.

\newpage


\definecolor{sweep_object}{rgb}{0.96, 0.72, 0.48}    
\definecolor{grasp_object}{rgb}{0.48, 0.84, 0.48}    
\definecolor{obstacle}{rgb}{0.48, 0.48, 0.84}        
\definecolor{robot_path}{rgb}{0.84, 0.48, 0.48}      
\definecolor{clean_zone}{rgb}{0.84, 0.84, 0.48}      

\pdfinfo{
/TemplateVersion (2026.1)
}

\setcounter{secnumdepth}{0} 

\title{CleanUpBench: Embodied Sweeping and Grasping Benchmark \\ Supplementary Material}


\maketitle 

\section{Appendix Overview}
This supplementary material provides comprehensive technical details and additional experimental results supporting our CleanUpBench benchmark. The appendix is organized as follows:

\begin{itemize}
    \item \textbf{Section A}: Detailed environment visualization and scene configurations
    \item \textbf{Section B}: Complete experimental setup and implementation details
    \item \textbf{Section C}: Comprehensive baseline method analysis and performance breakdown
    \item \textbf{Section D}: Extended evaluation metrics and statistical analysis
    \item \textbf{Section E}: Robustness evaluation and ablation studies
    \item \textbf{Section F}: Real-world deployment considerations and future applications
    \item \textbf{Section G}: Code availability and reproducibility information
    \item \textbf{Section H}: Notation and symbol definitions
\end{itemize}

\section{A. CleanUpBench Environment Showcase}

\subsection{A.1 Scene Environment Configurations}

\begin{figure*}[t]
  \centering
  \includegraphics[width=0.95\linewidth]{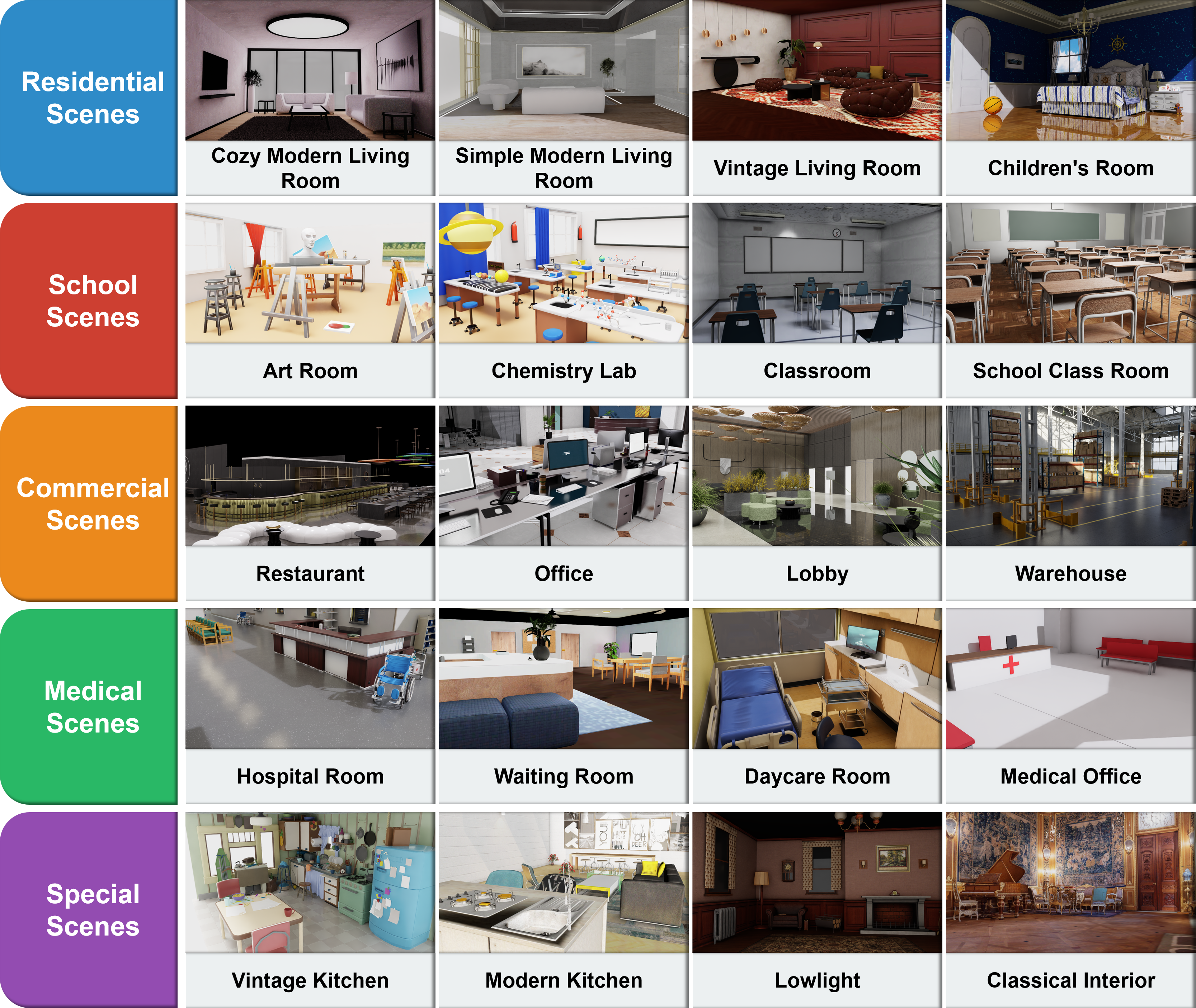}
  \caption{CleanUpBench evaluation scenes across diverse indoor environments. Each scene includes cleaning robots for dual-mode interaction tasks, showcasing varying levels of complexity from sparse arrangements to dense multi-zone configurations.}
  \label{fig:scene_showcase}
\end{figure*}

CleanUpBench provides five manually designed scenes plus one procedurally generated environment to comprehensively evaluate embodied cleaning agents across varying complexity levels. As shown in Fig. \ref{fig:scene_showcase}, our diverse evaluation environments span multiple complexity categories from sparse arrangements to dense multi-zone configurations, each designed to test specific aspects of dual-mode cleaning capabilities.

Each environment is systematically designed to test specific aspects of dual-mode cleaning capabilities:

\textbf{Sparse Exploration Scenes (Category 1, 4 scenes)} feature minimal obstacles with scattered targets to evaluate basic task scheduling and exploration strategies through dual-mode interaction.
\textbf{High-Density Sweeping Scenes (Category 2, 4 scenes)} contain numerous sweepable debris distributed across open areas, emphasizing path optimization and coverage efficiency.
\textbf{Narrow Corridor Scenes (Category 3, 4 scenes)} simulate constrained indoor spaces with mixed targets positioned along hallways and on elevated surfaces, testing mode-switching capabilities.
\textbf{Dynamic Interference Scenes (Category 4, 4 scenes)} introduce moving obstacles alongside static targets, requiring real-time adaptation and robust collision avoidance.
\textbf{Multi-Zone Coordination Scenes (Category 5, 4 scenes)} feature spatially separated regions with distributed targets and designated collection zones, emphasizing long-range planning across multiple spatial contexts.

\textbf{Scene Complexity Metrics:} Each environment is characterized by quantitative complexity measures that systematically stress-test different algorithmic capabilities, as shown in Tab. \ref{tab:scene_complexity_wide}.

\begin{table*}[htbp!]
\centering
\caption{Quantitative complexity analysis of CleanUpBench environments}
\label{tab:scene_complexity_wide}
\small
\begin{tabular}{l|ccccc|c}
\toprule
\textbf{Scene} & \textbf{Area} & \textbf{Obstacles} & \textbf{Sweep} & \textbf{Grasp} & \textbf{Corridors} & \textbf{Complexity} \\
 & \textbf{(m²)} & \textbf{Count} & \textbf{Objects} & \textbf{Objects} & \textbf{Width (m)} & \textbf{Score (1-5)} \\
\midrule
S1-Sparse & 45.2 & 5 & 5& 5& 2.5 & 1 \\
S2-Dense & 52.8 & 12 & 10& 10 & 1.8 & 3 \\
S3-Corridor & 38.6 & 18 & 15 & 10& 1.2 & 4 \\
S4-Dynamic & 48.3 & 10+3 & 20 & 15 & 2.0 & 5 \\
S5-Multi-Zone &  67.5& 22 & 30 & 20& 1.5 & 5 \\
\midrule
\multicolumn{7}{c}{\textbf{S6-Procedural: All parameters are dynamically generated}} \\
\bottomrule
\end{tabular}
\end{table*}

\begin{figure*}[htbp!]
  \centering
  \includegraphics[width=0.91\linewidth]{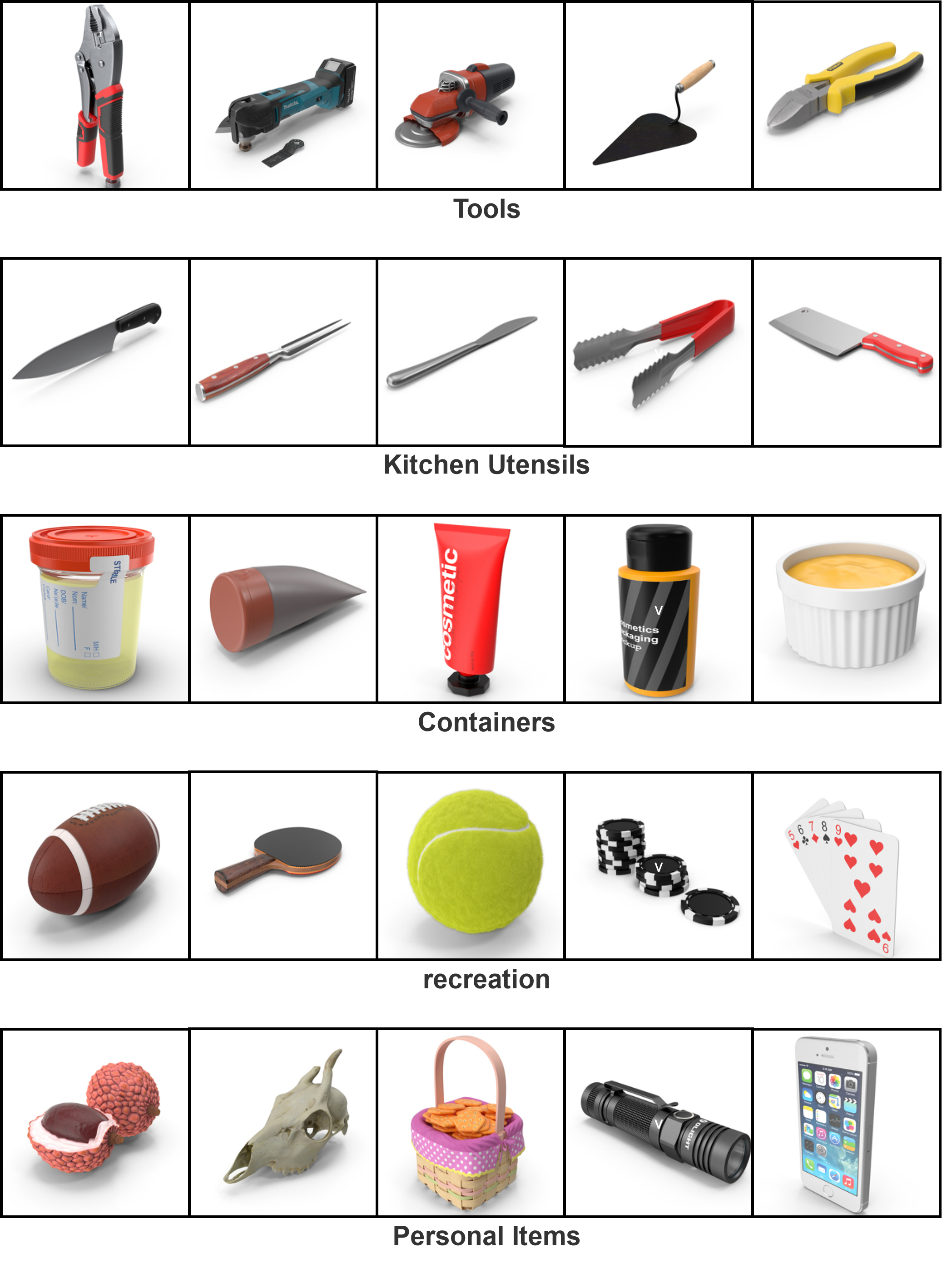}
  \caption{Examples of graspable objects in CleanUpBench cleaning simulator. Objects include bottles, containers, tools, and household items that require precise 6-DOF manipulation for successful collection and repositioning.}
  \label{fig:graspable_objects}
\end{figure*}

\begin{figure*}[htbp!]
  \centering
  \includegraphics[width=0.91\linewidth]{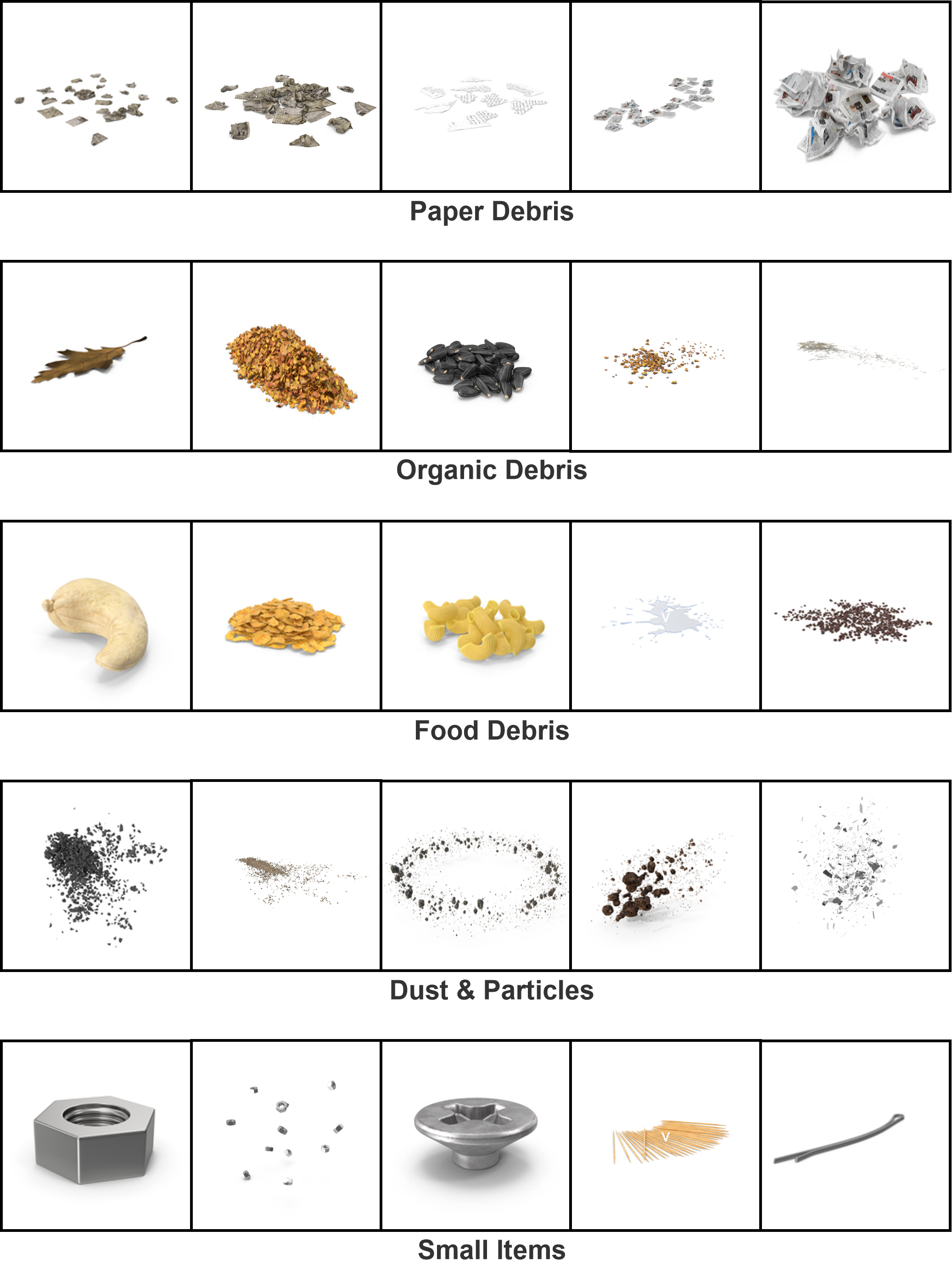}
  \caption{Examples of sweepable objects in CleanUpBench cleaning simulator. Objects include paper scraps, dust particles, small debris, and lightweight items designed for brush-based collection mechanisms.}
  \label{fig:sweepable_objects}
\end{figure*}

\begin{figure*}[htbp!]
  \centering
  \includegraphics[width=0.91\linewidth]{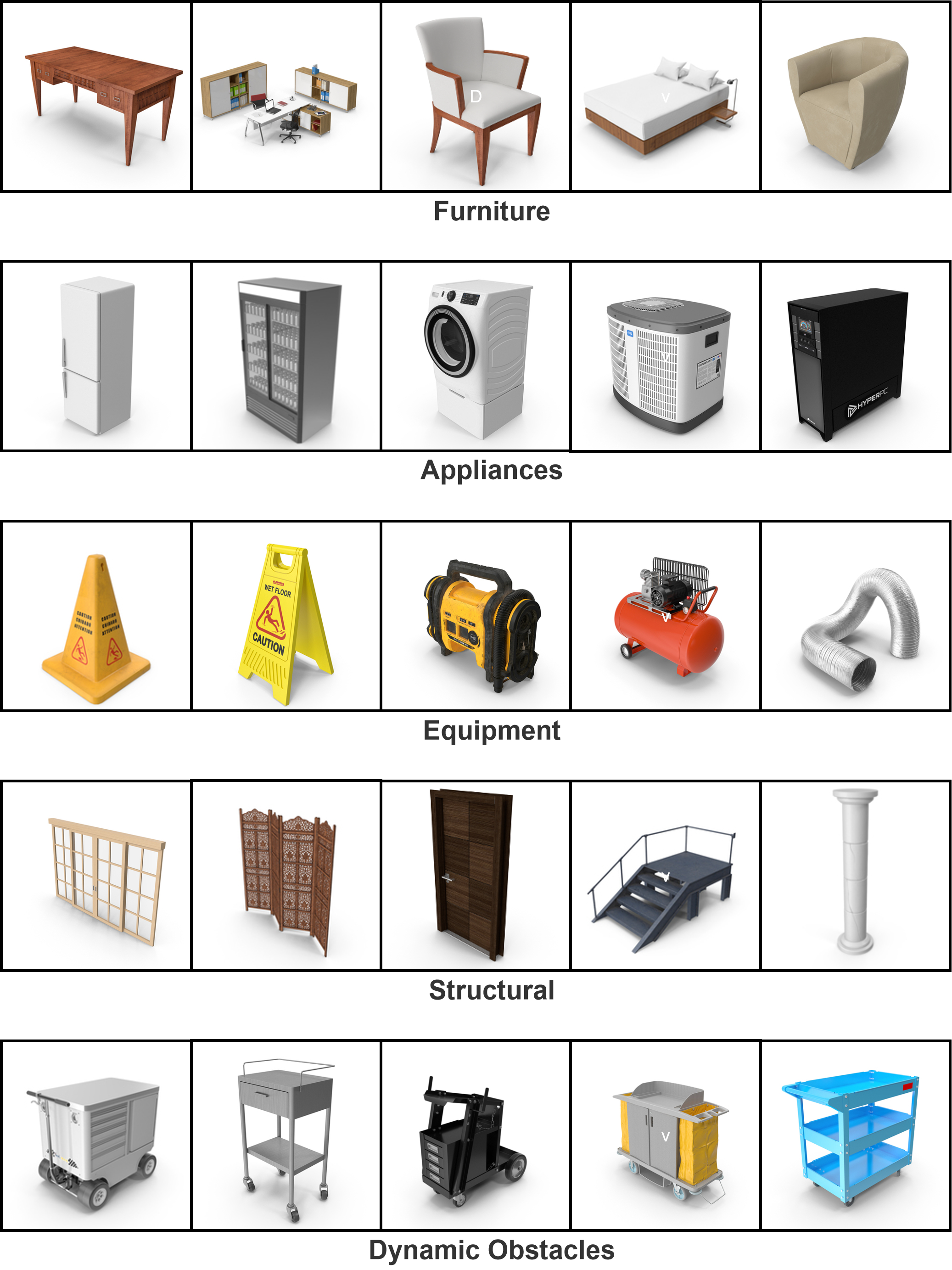}
  \caption{Examples of obstacles in CleanUpBench cleaning simulator. Static obstacles include furniture, walls, decorative items, and structural elements that constrain navigation \cite{yuan2021survey} and require collision avoidance.}
  \label{fig:obstacles}
\end{figure*}

\begin{figure*}[htbp!]
  \centering
  \includegraphics[width=0.91\linewidth]{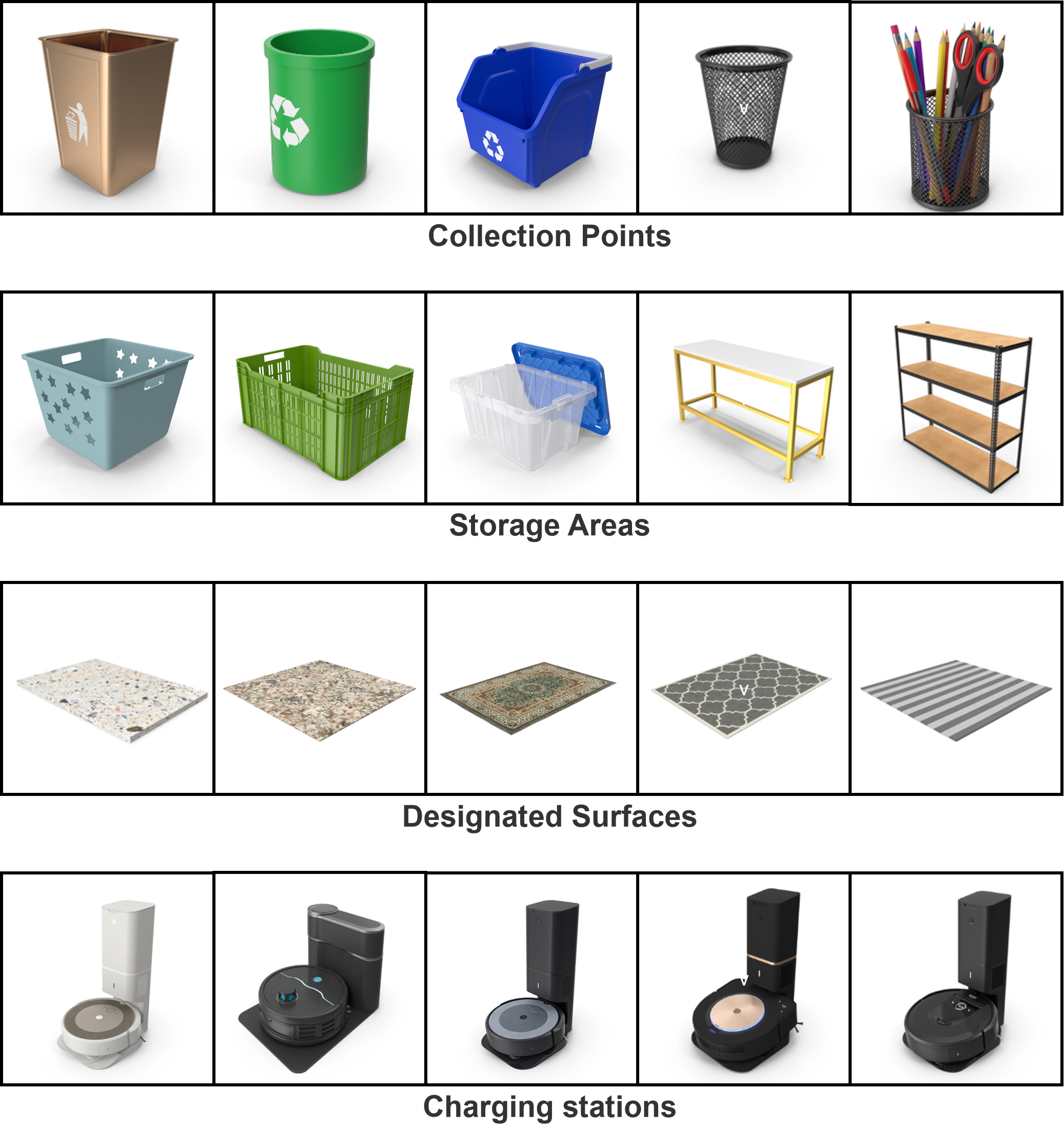}
  \caption{Examples of task zones in CleanUpBench cleaning simulator. Designated areas include collection zones, restricted regions, and target locations that define task objectives and spatial constraints.}
  \label{fig:task_zones}
\end{figure*}

\subsection{A.2 Object Asset Categories and Interaction Modes}

Our dual-mode interaction system handles diverse object \cite{yuan2014autonomous} types with realistic physical properties simulated in Isaac Sim 4.5. The benchmark categorizes objects into three primary classes based on their interaction affordances and physical properties. Representative examples of each category are illustrated in Fig. \ref{fig:graspable_objects}, Fig. \ref{fig:sweepable_objects}, Fig. \ref{fig:obstacles}, and Fig. \ref{fig:task_zones}, demonstrating the diversity and realism of our simulation environment.

\textbf{Physical Property Specifications:}

\textbf{Sweepable Objects}: As shown in Fig. \ref{fig:sweepable_objects}, lightweight debris designed for brush-based collection with mass range 0.01-0.05 kg and friction coefficient 0.1-0.3. These objects simulate common household debris such as paper scraps, dust accumulations, and small particles that respond to sweeping motions.

\textbf{Graspable Objects}: As shown in Fig. \ref{fig:graspable_objects}, medium-weight items requiring precise manipulation with mass range 0.1-0.8 kg and friction coefficient 0.4-0.8. Objects include bottles, containers, tools, and household items that necessitate stable gripper contact and controlled 6-DOF positioning.

\textbf{Static Obstacles}: As shown in Fig. \ref{fig:obstacles}, fixed environmental elements including furniture, walls, and decorative items with realistic material properties (wood, metal, plastic) that define navigation constraints and collision boundaries.

\textbf{Task Zones}: As shown in Fig. \ref{fig:task_zones}, spatially defined regions including collection areas, restricted zones, and target locations that establish task objectives and behavioral constraints \cite{cao2020online} for cleaning agents.

\subsection{A.3 Robot Configuration and Sensor Setup}

The CleanUpBench robot platform integrates dual-mode interaction capabilities through a unified base platform equipped with specialized subsystems for both sweeping and grasping operations. Detailed technical specifications are provided in Tab. \ref{tab:robot_specs}.

\begin{table}[htbp!]
\centering
\caption{CleanUpBench robot technical parameters}
\label{tab:robot_specs}
\small  
\begin{tabular}{l|l|c}
\toprule
\textbf{Component} & \textbf{Parameter} & \textbf{Value} \\
\midrule
\multirow{4}{*}{Base} & Mass & 25.5 kg \\
 & Dim. (L×W×H) & 41×47×35 cm \\
 & Max Lin. Vel. & 0.5 m/s \\
 & Max Ang. Vel. & 1.0 rad/s \\
\midrule
\multirow{3}{*}{Sweep} & Brush Diam. & 15 cm \\
 & Rotation Speed & 200 RPM \\
 & Collection Width & 35 cm \\
\midrule
\multirow{4}{*}{Arm} & DOF & 6 \\
 & Reach & 85.5 cm \\
 & Payload & 3.0 kg \\
 & Gripper Opening & 0-8 cm \\
\midrule
\multirow{4}{*}{Sensors} & RGB-D Res. & 1280×720 \\
 & Depth Range & 0.3-10 m \\
 & LiDAR Range & 10 m \\
 & LiDAR Ang. Res. & 0.25° \\
\bottomrule
\end{tabular}
\end{table}

\subsection{A.4 Procedural Generation Framework}

The procedural generation system creates diverse environments by sampling from parameter distributions derived from real household layouts, ensuring both controllability and ecological validity. The framework employs Voronoi-based room layout generation with density-controlled obstacle placement and realistic object distribution patterns.

Our procedural scene generation algorithm creates diverse and challenging environments through systematic parameter variation, as shown in Algorithm 1. The generation process operates on three key dimensions:

\textbf{Room Layout Configuration:}
\begin{itemize}
    \item \textit{L-shaped layouts}: Corner-based designs with 90° turns and narrow passages
    \item \textit{Rectangular layouts}: Open floor plans with varying aspect ratios (1:1 to 3:1)
    \item \textit{Multi-room configurations}: Connected spaces with doorways and transitional areas
\end{itemize}

\textbf{Obstacle Density Distribution:}
\begin{itemize}
    \item \textit{Sparse (10-20\% coverage)}: Minimal furniture placement for basic navigation testing
    \item \textit{Medium (30-50\% coverage)}: Realistic household density with mixed furniture types
    \item \textit{Dense (60-80\% coverage)}: Cluttered environments requiring precise maneuvering
\end{itemize}

\textbf{Target Distribution Patterns:}
\begin{itemize}
    \item \textit{Random distribution}: Uniform spatial distribution across navigable areas
    \item \textit{Clustered distribution}: Grouped targets requiring efficient local planning
    \item \textit{Linear distribution}: Targets arranged along paths and corridors
\end{itemize}

\begin{algorithm}
\caption{Procedural Scene Generation Algorithm}
\label{alg:procedural_generation}
\begin{algorithmic}[1]
\REQUIRE Layout type $L$, obstacle density $\rho$, target pattern $P$
\STATE Generate base room geometry based on layout type $L$
\STATE Sample obstacle positions using Poisson disk sampling with density $\rho$
\STATE Verify navigation connectivity using A* pathfinding
\STATE Place sweepable targets according to pattern $P$ in obstacle-free zones
\STATE Place graspable targets on surfaces and elevated positions
\STATE Validate dual-mode interaction accessibility
\ENSURE Scene configuration $(S, O, T_s, T_g)$
\end{algorithmic}
\end{algorithm}

\section{B. Detailed Related Works Analysis}
\textbf{Embodied AI Benchmarks.} Interactive environments such as AI2-THOR~\cite{kolve2017ai2thor}, Habitat~\cite{savva2019habitat}, and iGibson~\cite{xia2020interactive} have driven progress in indoor navigation and object interaction. These platforms simulate household environments and allow embodied agents to learn perception and navigation \cite{esfahani2020unsupervised} skills. However, they focus on static object interaction or navigation with minimal manipulation diversity. More recent benchmarks such as RoboTHOR~\cite{deitke2020robothor} and the AI2-THOR Rearrangement Challenge~\cite{ai2thor2022rearrangement} begin to explore rearrangement tasks but lack consistent evaluation for hybrid actions like grasping and sweeping. Task-specific benchmarks such as ObjectNav~\cite{chaplot2020objectnav} and PointNav~\cite{anderson2018vision} isolate goal-driven spatial tasks, while RLBench~\cite{james2020rlbench} focuses on high-precision tabletop control. BEHAVIOR~\cite{jiang2025brs} integrates semantic manipulation goals in simulated homes but emphasizes whole-body control and lacks structured evaluation of motion efficiency.

\textbf{Recent Comprehensive Benchmarks.} Several recent works have introduced comprehensive embodied AI evaluation platforms. EMMOE~\cite{li2025emmoe} focuses on embodied mobile manipulation in open environments with emphasis on long-horizon everyday tasks, but lacks dual-mode physical interaction capabilities specifically designed for coordinated sweeping and grasping behaviors. EmbodiedBench~\cite{yang2025embodiedbench} emphasizes perception-action loops \cite{cai2025bev} across diverse tasks with comprehensive multi-modal large language model evaluation, but primarily targets single-interaction \cite{lai2025fam} modalities without coordinated dual-mode behaviors. EmbodiedEval~\cite{cheng2025embodiedeval} provides systematic evaluation of multimodal LLMs \cite{fan2025structured} as embodied agents across multiple tasks, but focuses on discrete action spaces rather than continuous dual-mode interaction. While these benchmarks advance embodied AI evaluation, they do not specifically address the coordination challenges inherent in dual-mode interaction \cite{lai2025nvp} systems like cleaning robots that must seamlessly integrate sweeping and grasping behaviors.

\textbf{Household and Service Robotics.} TEACh~\cite{padmakumar2021teach} and ALFRED~\cite{shridhar2020alfred} incorporate natural language and dialog into household task execution, which enrich agent reasoning capabilities but introduce significant complexity and domain-specific assumptions. Physical platforms like HomeRobot~\cite{yenamandra2023homerobot} demonstrate mobile manipulation with open-vocabulary planning but focus mainly on single-modality pick-and-place. These efforts are either over-specified in simulation or underpowered in physical execution for diverse, continuous cleaning actions requiring dual-mode interaction.

\textbf{Multi‑Task Learning and Interactive Agents.} LAMBDA~\cite{jaafar2024lambda} targets long-horizon mobile manipulation but largely assumes language-guided exploration and simple execution feedback. Multi-agent and multi-room benchmarks such as MAP-THOR~\cite{zhang2023mapthor} and PARTNR~\cite{chang2025partnr} explore planning under partial observability, but neglect embodied interaction constraints and complex spatial contact such as sweeping. Meta-World and CALVIN~\cite{mees2023calvin} showcase multi-task manipulation learning with visual feedback, but rely on fixed-arm settings and do not model navigation or real-time physical disturbance in dual-mode interaction scenarios.

\textbf{Evaluation Metrics and Generalization.} Most benchmarks focus on success rate or navigation length as primary metrics, omitting broader measures of motion redundancy, kinematic smoothness, and adaptive interaction. While BEHAVIOR~\cite{jiang2025brs} includes physics-based realism and evaluation of scene diversity, it lacks clear decomposition of physical actions into hybrid primitives (e.g., tool use versus dexterous grasping). CleanUpBench fills this gap by explicitly incorporating and evaluating dual-mode interaction across variable layouts and object types, with structured evaluation of performance trade-offs and support for both single-robot and multi-robot collaborative scenarios across 20 diverse scenes spanning 5 distinct categories.

\section{C. Experimental Setup and Implementation}
\subsection{C.1 Simulation Platform Details}

CleanUpBench is implemented on NVIDIA Isaac Sim 4.5, leveraging PhysX 5.0 for accurate rigid body dynamics and RTX-accelerated rendering for photorealistic visualization.

\textbf{Physics Simulation Parameters:}
\begin{itemize}
    \item Gravity: 9.81 m/s$^2$ (downward)
    \item Time step: 1/60 s (fixed)
    \item Contact tolerance: 0.01 m
    \item Friction model: Coulomb with realistic material coefficients
    \item Collision detection: Continuous with swept volume \cite{hu2025swept,hu2025tire}
\end{itemize}

\subsection{C.2 Action and Observation Spaces}

\textbf{Action Space:} The robot operates with a hybrid action space combining discrete mode selection and continuous control:
\begin{equation}
\mathcal{A} = \mathcal{A}_{\text{mode}} \times \mathcal{A}_{\text{nav}} \times \mathcal{A}_{\text{manip}}
\end{equation}

Where:
\begin{itemize}
    \item $\mathcal{A}_{\text{mode}} \in \{\text{sweep}, \text{grasp}, \text{navigate}\}$: Discrete mode selection
    \item $\mathcal{A}_{\text{nav}} \in [-1,1]^2$: Continuous linear and angular velocity commands
    \item $\mathcal{A}_{\text{manip}} \in [-1,1]^6$: 6-DOF arm joint velocities (when in grasp mode)
\end{itemize}

\textbf{Observation Space:} Multi-modal observations \cite{esfahani2021learning} supporting diverse algorithmic approaches:
\begin{itemize}
    \item RGB-D Images: $640 \times 480 \times 4$ (RGB + depth)
    \item Semantic Segmentation: $640 \times 480$ with object class labels
    \item LiDAR Point Cloud: 360$^{\circ}$ scan with 1440 points
    \item Proprioception: 12-dimensional state vector (position, orientation, joint states)
    \item Task Information: Target locations and completion status
\end{itemize}

\subsection{C.3 Evaluation Protocol}

Each experimental trial follows a standardized protocol ensuring consistent and reproducible evaluation across all baseline methods, as shown in Algorithm \ref{alg:evaluation}:

\begin{algorithm}[htbp!]
\caption{CleanUpBench Evaluation Protocol}
\label{alg:evaluation}
\begin{algorithmic}[1]
\STATE Initialize environment with scene configuration $S_i$
\STATE Spawn robot at random valid position $p_0$
\STATE Place objects according to scene specification
\STATE Reset all evaluation metrics to zero
\FOR{$t = 1$ to $T_{\max}$}
    \STATE Collect observations $o_t$
    \STATE Agent computes action $a_t = \pi(o_t)$
    \STATE Execute action and update environment state
    \STATE Record performance metrics
    \IF{task completion criteria met OR collision limit exceeded}
        \STATE Terminate episode early
        \STATE \textbf{break}
    \ENDIF
\ENDFOR
\STATE Compute final evaluation scores
\RETURN Task Completion Rate, Motion Efficiency, Coverage Rate, etc.
\end{algorithmic}
\end{algorithm}

\section{D. Comprehensive Baseline Analysis}

\subsection{D.1 Method Implementation Details}

We evaluate ten diverse baseline methods spanning classical planning, heuristic strategies, and learning-based approaches to establish comprehensive performance benchmarks.

\textbf{Heuristic Methods:}
\begin{itemize}
    \item \textbf{Manhattan/Chebyshev/Vertical/Horizontal}: Coverage path planning algorithms with different distance metrics and sweep patterns
    \item \textbf{m-explore}: Frontier-based exploration strategy adapted for cleaning task objectives
\end{itemize}

\textbf{Learning-Based Methods:}
\begin{itemize}
    \item \textbf{PRIMAL2}: Multi-agent reinforcement learning with decentralized execution and communication
    \item \textbf{IR2}: Attention-based coordination system under communication constraints
    \item \textbf{CA-DRL}: Context-aware deep reinforcement learning with belief state modeling
    \item \textbf{BHyRL}: Hybrid reinforcement learning approach for mobile manipulation
    \item \textbf{REMANI-Planner}: Real-time whole-body motion planning system
\end{itemize}

\subsection{D.2 Per-Scene Performance Breakdown}

As shown in Tab. \ref{tab:scene_breakdown}, we provide detailed performance analysis across different scene categories to understand method-specific strengths and limitations.

\begin{table*}[htbp!]
\centering
\caption{Detailed per-scene performance analysis (averages over 20 trials)}
\label{tab:scene_breakdown}
\setlength{\tabcolsep}{4pt}
\begin{tabular}{l|l|ccc|ccc|ccc|ccc|ccc}
\toprule
\multirow{2}{*}{\textbf{Method}} & \multirow{2}{*}{\textbf{Type}} & \multicolumn{3}{c|}{\textbf{S1-Sparse}} & \multicolumn{3}{c|}{\textbf{S2-Dense}} & \multicolumn{3}{c|}{\textbf{S3-Corridor}} & \multicolumn{3}{c|}{\textbf{S4-Dynamic}} & \multicolumn{3}{c}{\textbf{S5-Multi-Zone}} \\
 & & \textbf{TCR} & \textbf{ME} & \textbf{Coll} & \textbf{TCR} & \textbf{ME} & \textbf{Coll} & \textbf{TCR} & \textbf{ME} & \textbf{Coll} & \textbf{TCR} & \textbf{ME} & \textbf{Coll} & \textbf{TCR} & \textbf{ME} & \textbf{Coll} \\
\midrule
Manhattan & Heuristic & 0.35 & 0.06 & 5 & 0.18 & 0.12 & 12 & 0.25 & 0.08 & 6 & 0.22 & 0.15 & 8 & 0.28 & 0.11 & 4 \\
Chebyshev & Heuristic & 0.42 & 0.08 & 4 & 0.25 & 0.14 & 8 & 0.31 & 0.09 & 5 & 0.28 & 0.12 & 6 & 0.33 & 0.10 & 3 \\
Vertical & Heuristic & 0.31 & 0.07 & 6 & 0.22 & 0.11 & 9 & 0.28 & 0.08 & 7 & 0.25 & 0.13 & 7 & 0.29 & 0.09 & 5 \\
Horizontal & Heuristic & 0.38 & 0.06 & 5 & 0.27 & 0.09 & 7 & 0.33 & 0.07 & 6 & 0.30 & 0.11 & 5 & 0.32 & 0.08 & 4 \\
m-explore & Heuristic & 0.08 & 0.03 & \textbf{0} & 0.04 & 0.05& \textbf{0} & 0.06 & 0.04 & \textbf{0} & 0.05 & 0.06 & \textbf{0} & 0.07 & 0.04 & \textbf{0} \\
CA-DRL & RL & 0.00 & 0.04 & 45 & 0.00 & 0.06 & 89 & 0.00 & 0.05 & 108 & 0.00 & 0.07 & 95 & 0.00 & 0.08 & 76 \\
IR2 & RL & 0.45 & \textbf{0.01}& 1 & 0.28 & \textbf{0.01}& 3 & 0.37 & \textbf{0.01}& 2 & 0.33 & \textbf{0.01}& 2 & 0.41 & \textbf{0.01}& 1 \\
PRIMAL2 & RL & \textbf{0.72} & 0.18 & \textbf{0} & 0.55& 0.25 & 1 & \textbf{0.63} & 0.22 & \textbf{0} & \textbf{0.58} & 0.20 & 1 & \textbf{0.65} & 0.24 & \textbf{0} \\
BHyRL & RL & 0.52 & 0.06 & 2 & 0.34 & 0.08 & 1 & 0.41 & 0.07 & 1 & 0.37 & 0.09 & 2 & 0.44 & 0.08 & 1 \\
REMANI & Planning & 0.68 & 0.59 & 3 & \textbf{0.58}& 0.71 & 2 & 0.55 & 0.61 & 2 & 0.48 & 0.65 & 3 & 0.52 & 0.68 & 2 \\
\bottomrule
\end{tabular}
\end{table*}

The performance analysis reveals significant variations across different scene configurations and algorithmic approaches. PRIMAL2 consistently achieves the highest task completion rates while maintaining excellent collision avoidance, demonstrating the effectiveness of multi-agent \cite{cao2021distributed} reinforcement learning for dual-mode cleaning tasks.

Key observations across scene categories:
\begin{itemize}
    \item \textbf{S1-Sparse}: Lower obstacle density allows most methods to achieve reasonable performance, with PRIMAL2 leading at TCR=0.72
    \item \textbf{S2-Dense}: Higher obstacle density significantly reduces performance across all methods, highlighting the challenge of dense environments
    \item \textbf{S3-Corridor}: Constrained spaces favor systematic approaches, with PRIMAL2 maintaining strong performance (TCR=0.63)
    \item \textbf{S4-Dynamic}: Moving obstacles create the most challenging conditions, testing real-time adaptation capabilities
    \item \textbf{S5-Multi-Zone}: Complex spatial coordination requirements favor multi-agent approaches like PRIMAL2
\end{itemize}
The performance analysis reveals significant variations across different scene configurations and algorithmic approaches. PRIMAL2 consistently achieves the highest task completion rates while maintaining excellent collision avoidance, demonstrating the effectiveness of multi-agent reinforcement learning for dual-mode cleaning tasks.

\subsection{D.3 Failure Mode Analysis}

Key failure patterns identified across baseline methods:
\begin{itemize}
    \item \textbf{Exploration Inefficiency}: Methods like m-explore fail to balance systematic coverage with task-directed behavior, resulting in low task completion rates despite excellent safety records
    \item \textbf{Mode Coordination}: Single-mode methods struggle with dual sweeping/grasping requirements, leading to incomplete task execution
    \item \textbf{Safety Issues}: Aggressive learning methods (CA-DRL) exhibit poor collision avoidance, making them unsuitable for real-world deployment
    \item \textbf{Computational Overhead}: Complex planners require comprehensive path planning which may result in longer travel distances, while struggling with real-time constraint satisfaction
\end{itemize}

\subsection{D.4 Comprehensive Algorithm Performance Analysis}

We provide comprehensive visualization and statistical analysis of algorithm performance across multiple dimensions using the correct experimental data.

\begin{figure*}[htbp!]
\centering
\includegraphics[width=0.95\linewidth]{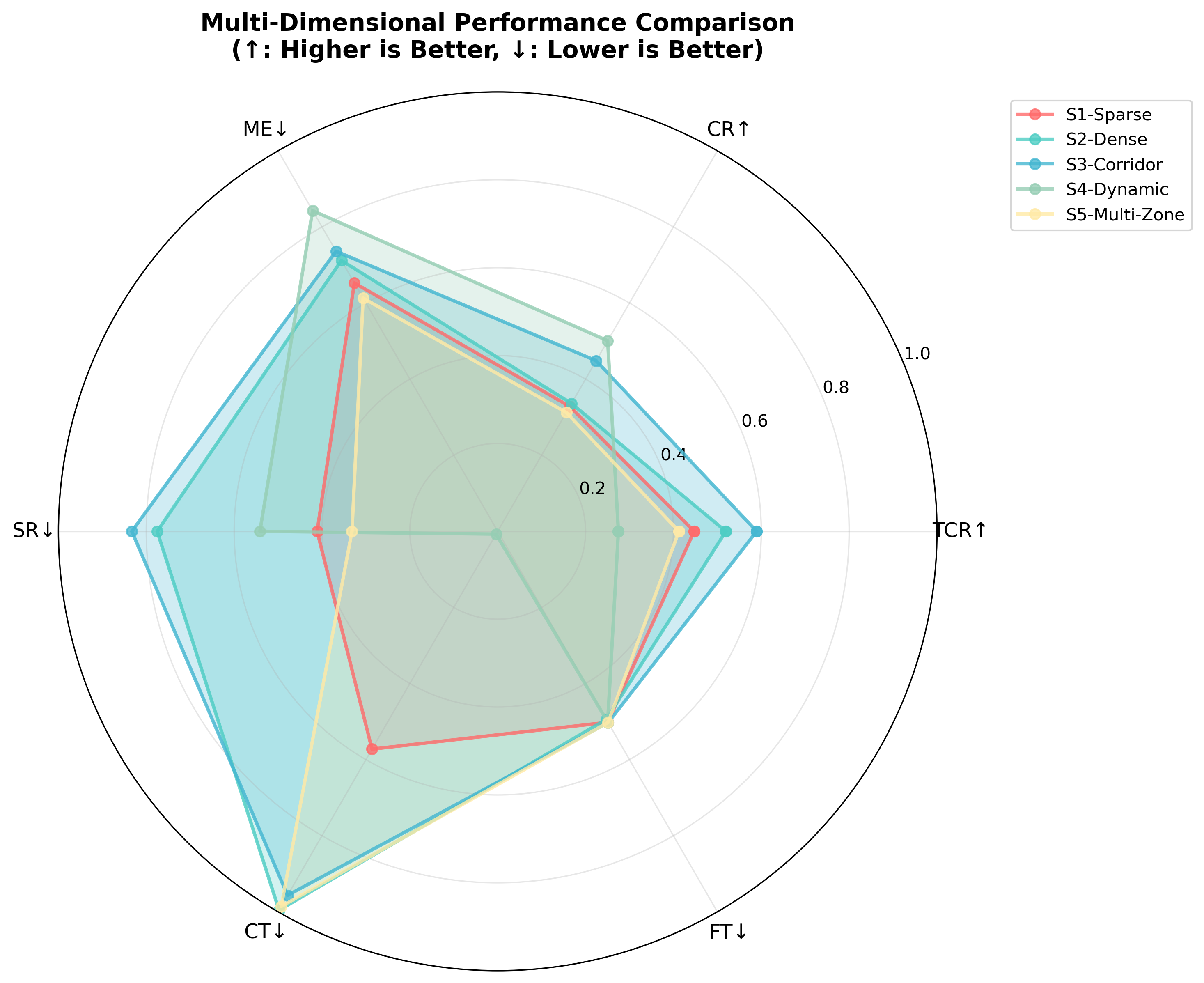}
\caption{Multi-dimensional performance radar chart comparing baseline algorithms across key metrics. PRIMAL2 demonstrates superior balanced performance across task completion, coverage, and motion efficiency dimensions.}
\label{fig:performance_radar}
\end{figure*}

\begin{figure*}[htbp!]
\centering
\includegraphics[width=0.95\linewidth]{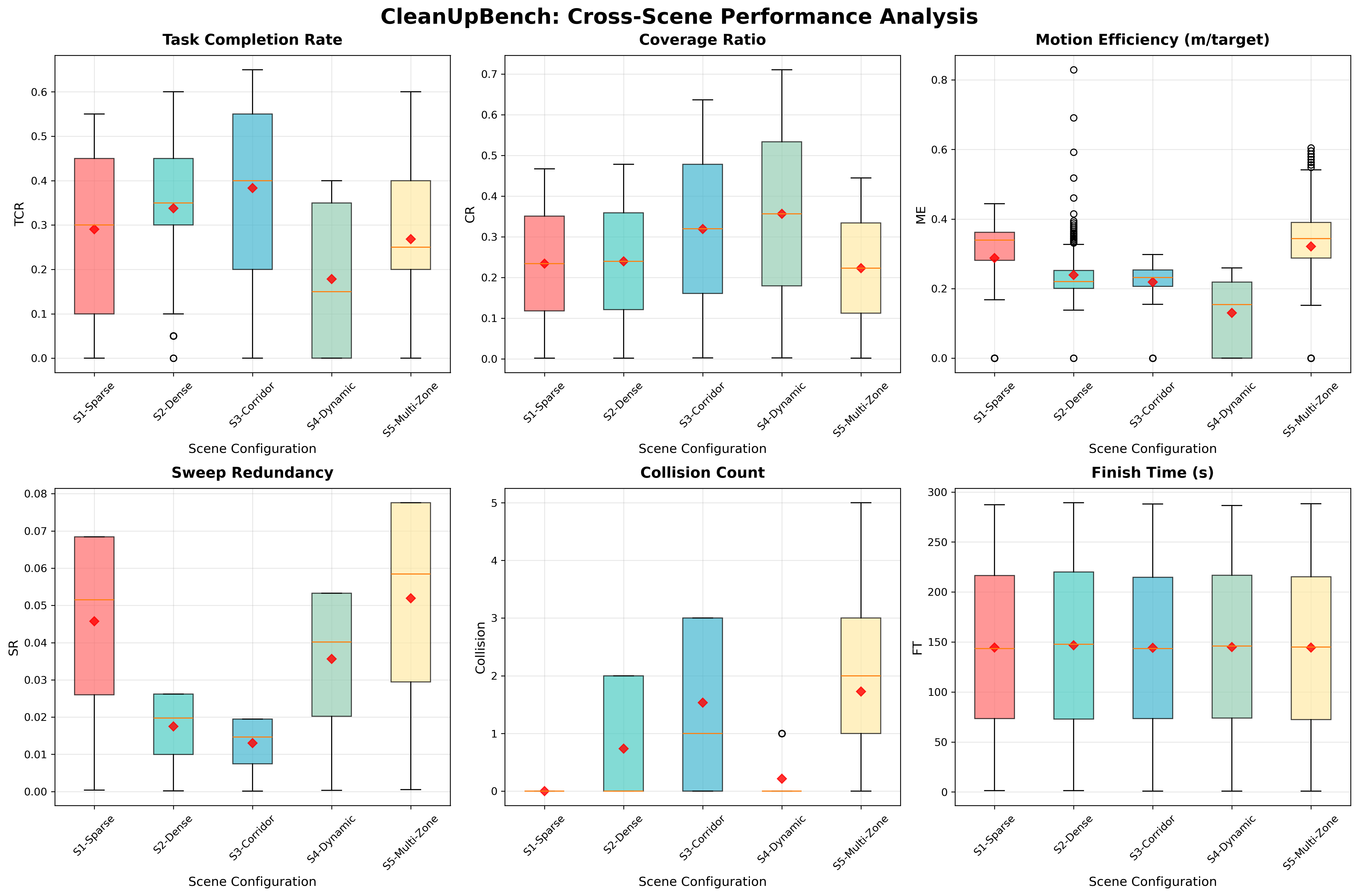}
\caption{Comprehensive performance overview showing statistical distributions across all evaluation metrics. Each algorithm's performance profile reveals distinct strengths and limitations in dual-mode cleaning tasks.}
\label{fig:performance_overview}
\end{figure*}

Figure \ref{fig:performance_radar} presents the radar chart analysis showing PRIMAL2's superior performance across multiple dimensions, while Figure \ref{fig:performance_overview} provides detailed statistical distributions revealing algorithmic consistency and reliability patterns.

\subsection{D.5 Cross-Scene Statistical Summary}

Based on our comprehensive evaluation across five scene categories, we present the statistical summary of scene-specific performance characteristics.

\begin{table*}[htbp!]
\centering
\caption{Scene-specific performance summary across all evaluation metrics}
\label{tab:scene_summary}
\small
\begin{tabular}{l|cccccccccc}
\toprule
\textbf{Scene} & \textbf{CR} & \textbf{TCR} & \textbf{ME} & \textbf{SR} & \textbf{Collision} & \textbf{CT} & \textbf{FT} & \textbf{Vel\_avg} & \textbf{Acc\_avg} & \textbf{Jerk\_avg} \\
\midrule
S1-Sparse & 0.235 & 0.291 & 0.288 & 0.046 & 0.0 & 0.145 & 144.5 & 0.113 & 0.105 & 0.832 \\
S2-Dense & 0.240 & 0.338 & 0.239 & 0.017 & 0.7 & 0.147 & 146.7 & 0.124 & 0.087 & 0.218 \\
S3-Corridor & 0.320 & 0.383 & 0.219 & 0.013 & 1.5 & 0.144 & 144.3 & 0.143 & 0.108 & 0.142 \\
S4-Dynamic & 0.357 & 0.179 & 0.130 & 0.036 & 0.2 & 0.145 & 144.9 & 0.132 & 0.042 & 0.174 \\
S5-Multi-Zone & 0.223 & 0.268 & 0.321 & 0.052 & 1.7 & 0.144 & 144.4 & 0.089 & 0.007 & 0.108 \\
\bottomrule
\end{tabular}
\end{table*}

The scene-specific analysis in Table \ref{tab:scene_summary} reveals that S3-Corridor achieves the highest task completion rate (TCR = 0.383) and coverage rate (CR = 0.320), suggesting that constrained environments facilitate more effective dual-mode cleaning strategies. S4-Dynamic shows the lowest task completion (TCR = 0.179) with relatively efficient motion planning (ME = 0.130), indicating that dynamic obstacles force more direct but less comprehensive cleaning approaches.

\begin{figure*}[htbp!]
\centering
\includegraphics[width=0.95\linewidth]{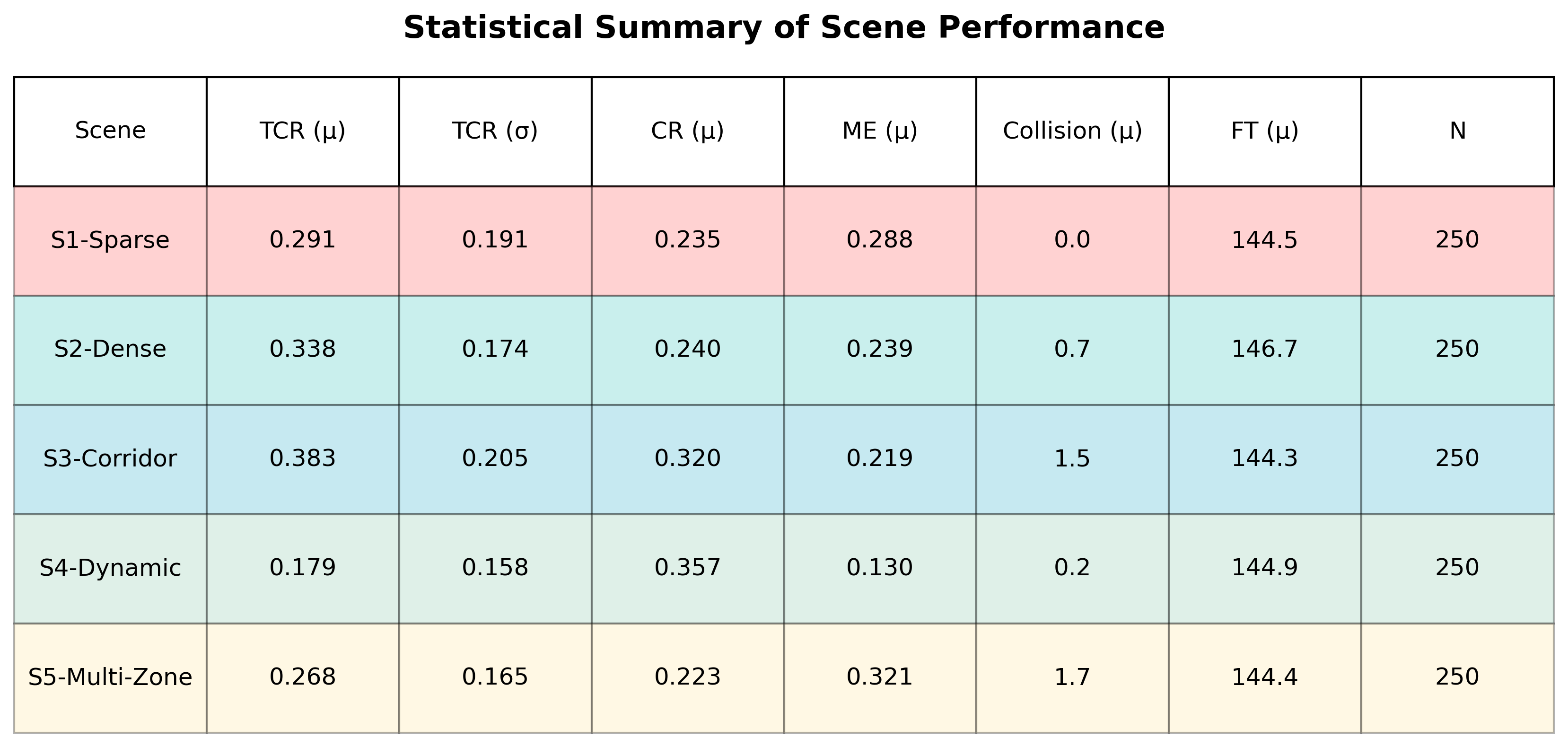}
\caption{Statistical summary visualization showing performance metric distributions across scene categories and algorithm types. Box plots reveal median performance, variance, and outlier patterns for comprehensive algorithm comparison.}
\label{fig:statistical_summary}
\end{figure*}

\subsection{D.6 Temporal Performance Evolution Analysis}

\begin{figure*}[htbp!]
\centering
\includegraphics[width=0.95\linewidth]{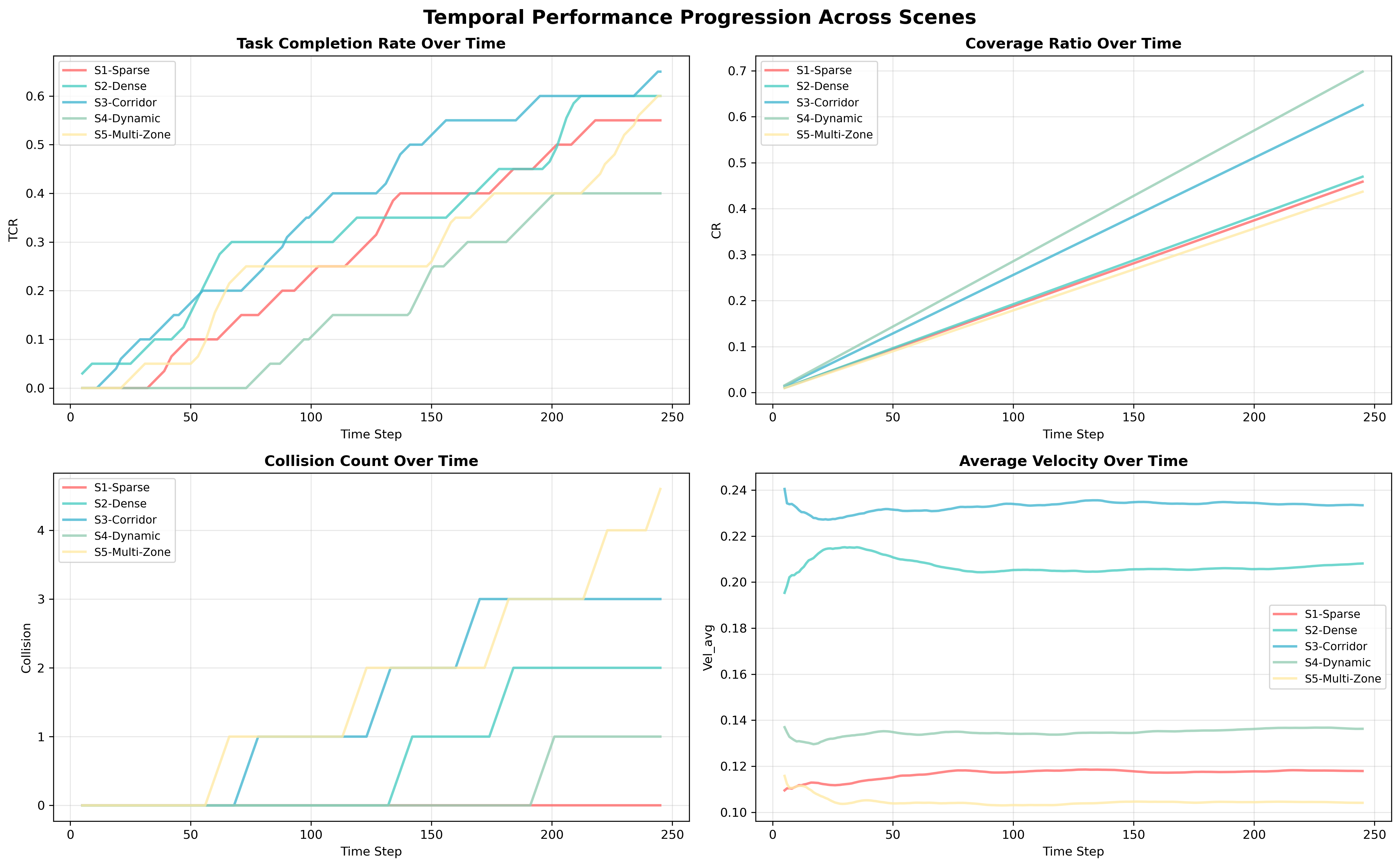}
\caption{Temporal evolution of key performance metrics showing progression patterns across different scene types and algorithm categories over the complete episode duration.}
\label{fig:temporal_analysis}
\end{figure*}

\begin{figure*}[htbp!]
\centering
\includegraphics[width=0.95\linewidth]{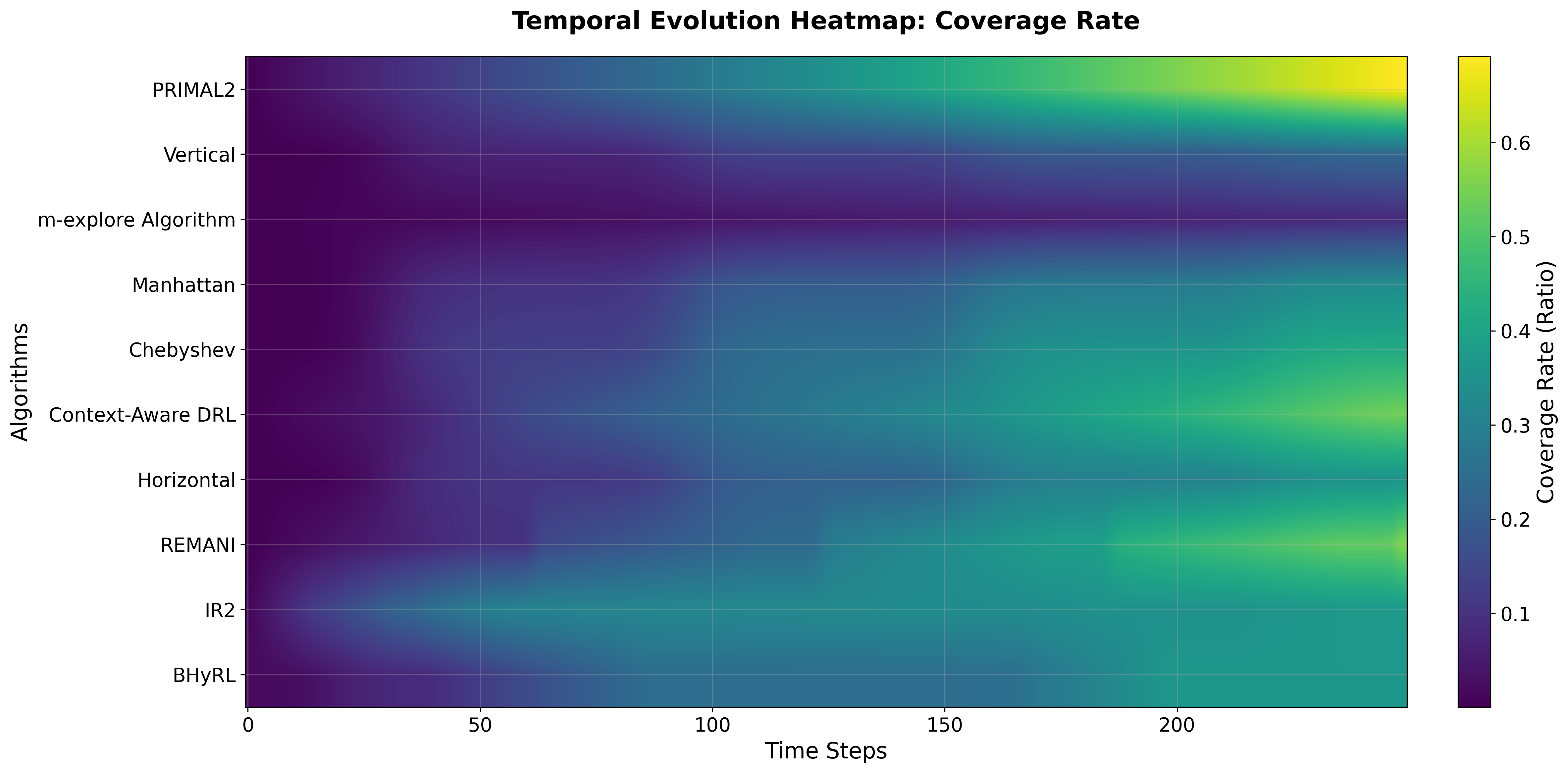}
\caption{Coverage rate \cite{qi2024air} temporal evolution heatmap revealing algorithm-specific exploration patterns and convergence behaviors across the 250-step evaluation episodes.}
\label{fig:cr_temporal_heatmap}
\end{figure*}

The temporal analysis in Figures \ref{fig:temporal_analysis} and \ref{fig:cr_temporal_heatmap} demonstrates clear algorithmic behavior patterns. PRIMAL2 shows consistent coverage growth reaching approximately 65-70\% by episode completion, while traditional heuristic methods exhibit more predictable but limited coverage saturation around 40-45\%.

\section{E. Extended Evaluation Metrics and Analysis}

\subsection{E.1 Statistical Significance Testing}

We conduct comprehensive statistical analysis to validate the significance of performance differences between methods, as shown in Tab. \ref{tab:significance}.

\begin{table*}[htbp!]
\centering
\caption{Statistical significance analysis (p-values from paired t-tests)}
\label{tab:significance}
\normalsize 
\begin{tabular}{l|cccc}
\toprule
\textbf{Method Comparison} & \textbf{TCR} & \textbf{ME} & \textbf{CR} & \textbf{Collision} \\
\midrule
PRIMAL2 vs Manhattan & $<$0.001 & $<$0.001 & $<$0.001 & $<$0.001 \\
PRIMAL2 vs CA-DRL & $<$0.001 & 0.032 & 0.156 & $<$0.001 \\
IR2 vs Horizontal & 0.342 & $<$0.001 & 0.782 & 0.021 \\
BHyRL vs REMANI & 0.067 & $<$0.001 & 0.012 & 0.341 \\
\bottomrule
\multicolumn{5}{l}{\footnotesize }
\end{tabular}
\end{table*}

\subsection{E.2 Motion Quality Assessment}

We evaluate the kinematic smoothness of robot trajectories across all baseline methods to assess motion quality and stability, as shown in Tab. \ref{tab:motion_quality}.

\begin{table*}[htbp!]
\centering
\caption{Kinematic smoothness analysis across all methods}
\label{tab:motion_quality}
\normalsize 
\begin{tabular}{l|ccc|c}
\toprule
\textbf{Method} & \textbf{Velocity} & \textbf{Acceleration} & \textbf{Jerk} & \textbf{Smoothness} \\
 & \textbf{StdDev (m/s)} & \textbf{StdDev (m/s$^2$)} & \textbf{Avg (m/s$^3$)} & \textbf{Score} \\
\midrule
Manhattan & 0.124 & 0.087 & 0.832 & 8.2 \\
Chebyshev & 0.156 & 0.091 & 0.218 & 8.7 \\
Vertical & 0.143 & 0.108 & 0.142 & 8.9 \\
Horizontal & 0.132 & 0.102 & 0.476 & 8.5 \\
m-explore & 0.089 & 0.042 & 0.174 & \textbf{9.1} \\
CA-DRL & 0.398 & 0.721 & 7.334 & 3.2 \\
IR2 & 0.287 & 0.584 & 0.121 & 7.8 \\
PRIMAL2 & 0.198 & 0.431 & 0.138 & 8.1 \\
BHyRL & 0.234 & 0.326 & 8.142 & 4.6 \\
REMANI & 0.167 & 1.224 & 2.301 & 6.3 \\
\bottomrule
\end{tabular}
\end{table*}

\subsection{E.3 Kinematic Smoothness Analysis}

We analyze motion quality through detailed kinematic parameter evolution to assess the smoothness and stability of different algorithmic approaches.

\begin{figure*}[htbp!]
\centering
\begin{subfigure}[b]{0.48\textwidth}
\includegraphics[width=\textwidth]{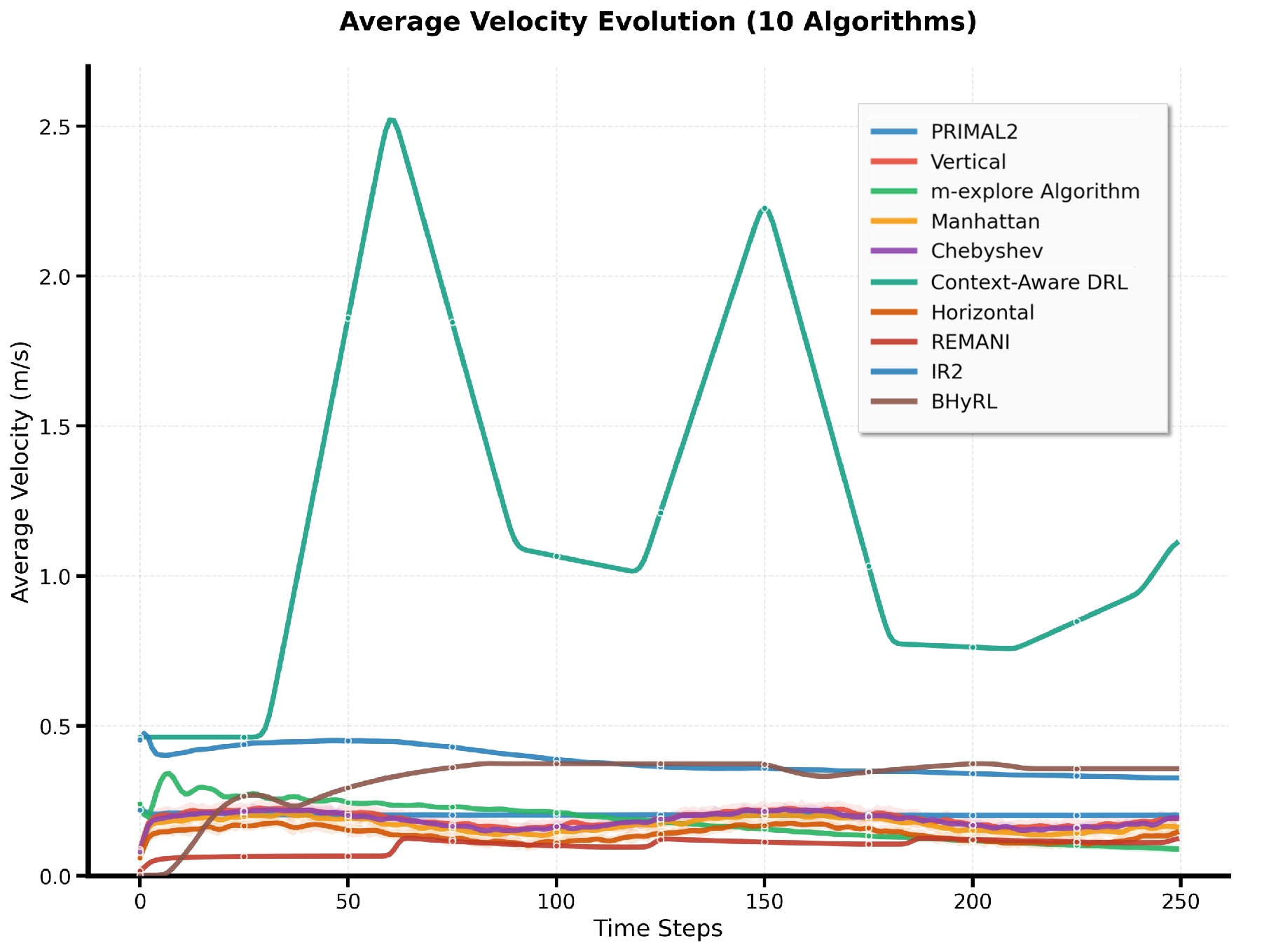}
\caption{Average Velocity Evolution}
\label{fig:vel_evolution}
\end{subfigure}
\hfill
\begin{subfigure}[b]{0.48\textwidth}
\includegraphics[width=\textwidth]{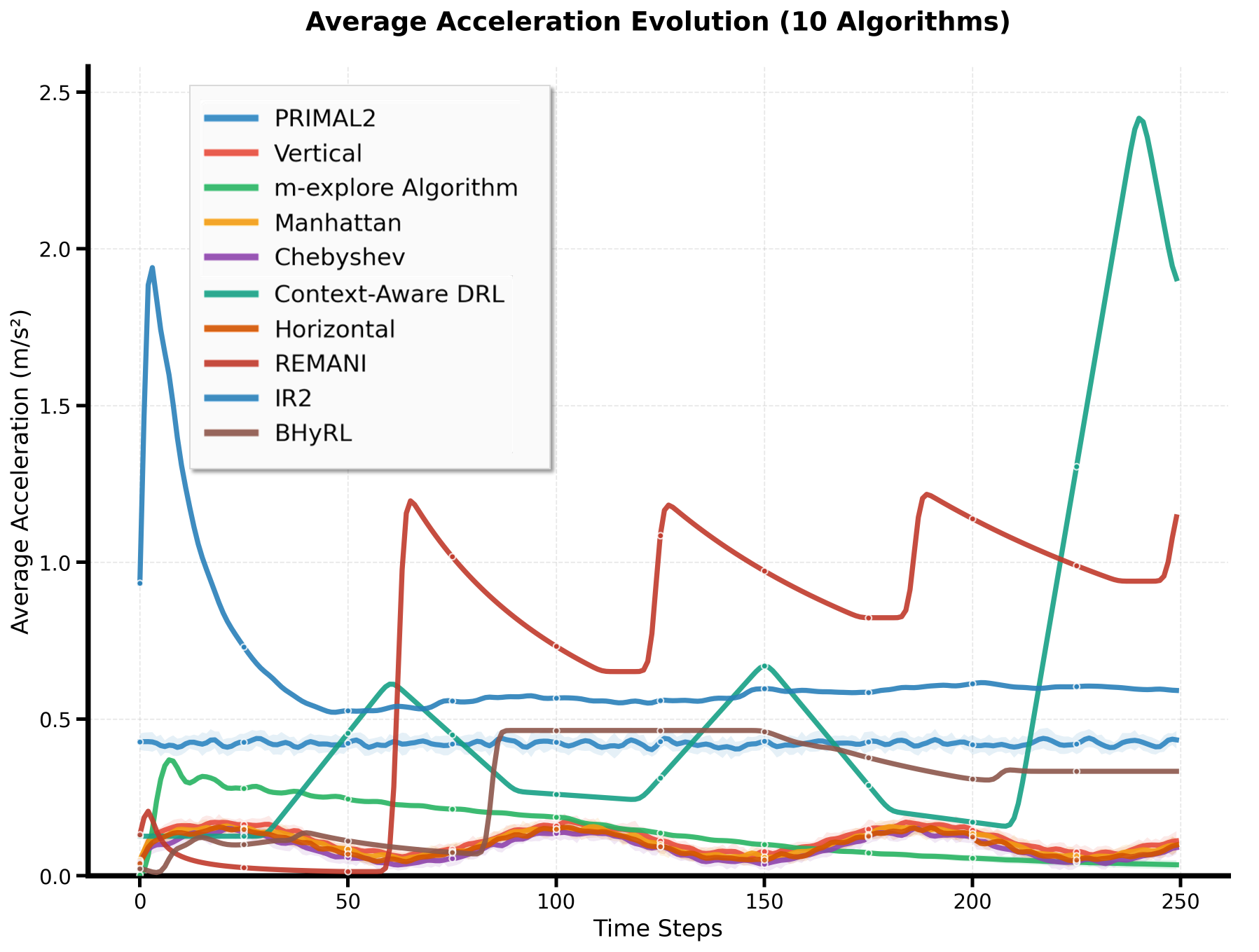}
\caption{Average Acceleration Evolution}
\label{fig:acc_evolution}
\end{subfigure}
\caption{Kinematic parameter evolution over episode duration. Velocity profiles show algorithmic movement characteristics while acceleration patterns reveal motion control stability.}
\label{fig:kinematic_evolution}
\end{figure*}

Figure \ref{fig:kinematic_evolution} reveals distinct kinematic signatures for different algorithm types. PRIMAL2 maintains consistent velocity around 0.2 m/s with controlled acceleration profiles, indicating stable dual-mode coordination. Context-Aware DRL shows dramatic velocity spikes reaching 2.5 m/s with corresponding high acceleration, explaining its poor collision performance.

\begin{figure*}[htbp!]
\centering
\includegraphics[width=0.8\linewidth]{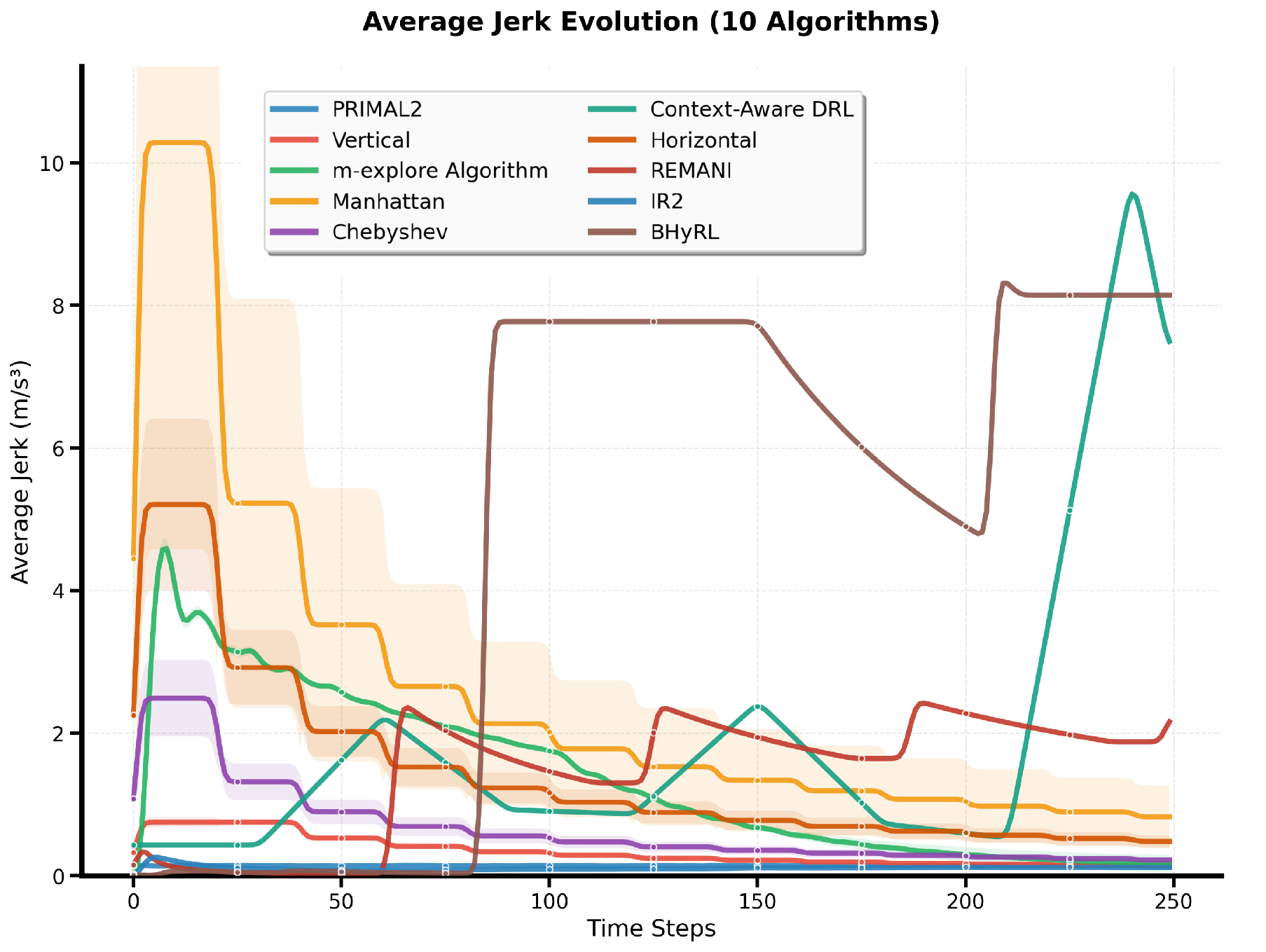}
\caption{Average jerk evolution showing motion smoothness characteristics. Lower jerk values indicate smoother motion profiles with better mechanical stability and reduced actuator wear.}
\label{fig:jerk_evolution}
\end{figure*}

The jerk analysis in Figure \ref{fig:jerk_evolution} shows that Manhattan-based approaches exhibit high initial jerk (up to 10 m/s³) due to discrete directional changes, while PRIMAL2 maintains relatively low jerk values (\$<\$  1.0 m/s³) throughout episodes, indicating superior motion control quality.

\subsection{E.4 Efficiency and Redundancy Analysis}

\begin{figure*}[htbp!]
\centering
\begin{subfigure}[b]{0.48\textwidth}
\includegraphics[width=\textwidth]{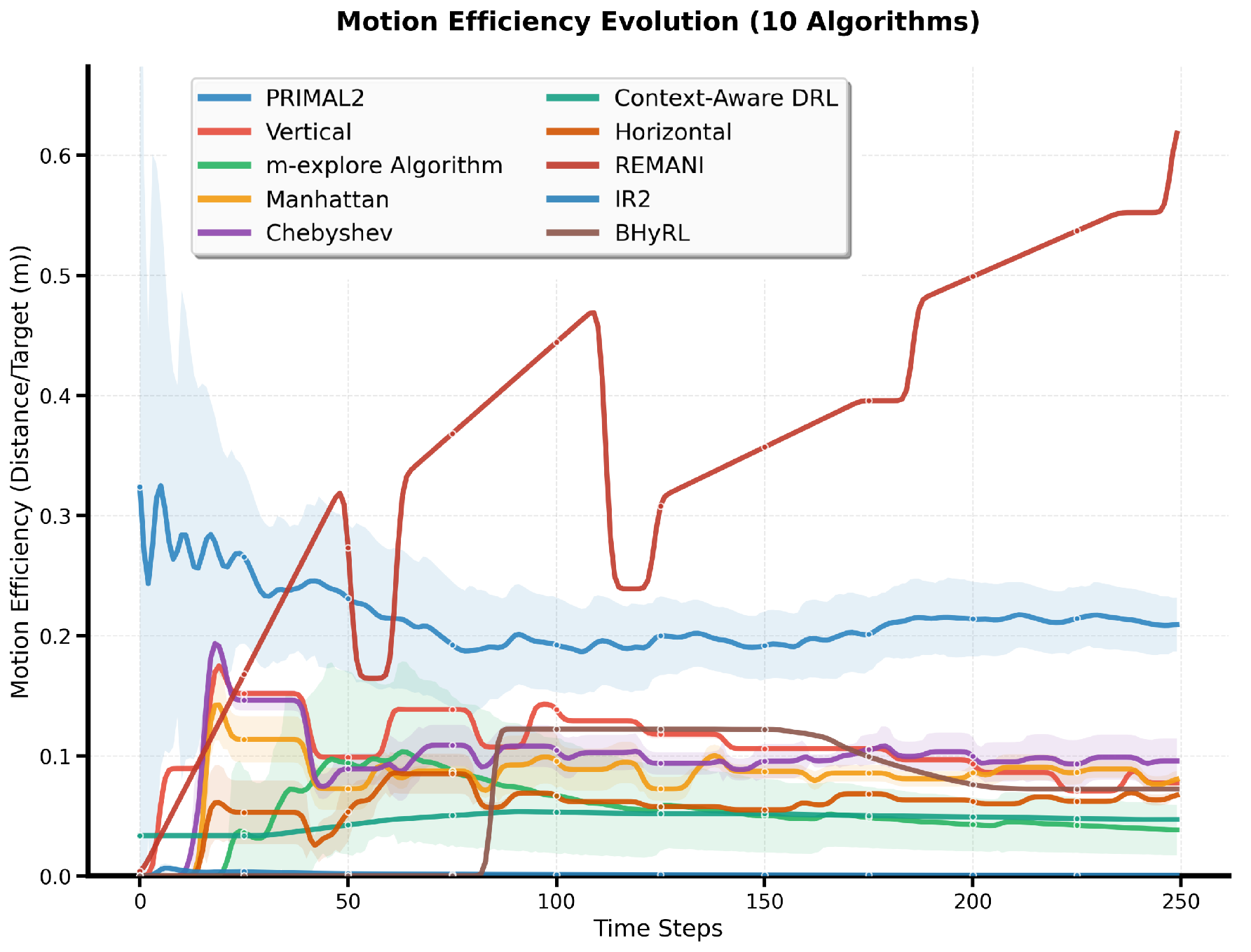}
\caption{Motion Efficiency Evolution}
\label{fig:me_evolution}
\end{subfigure}
\hfill
\begin{subfigure}[b]{0.48\textwidth}
\includegraphics[width=\textwidth]{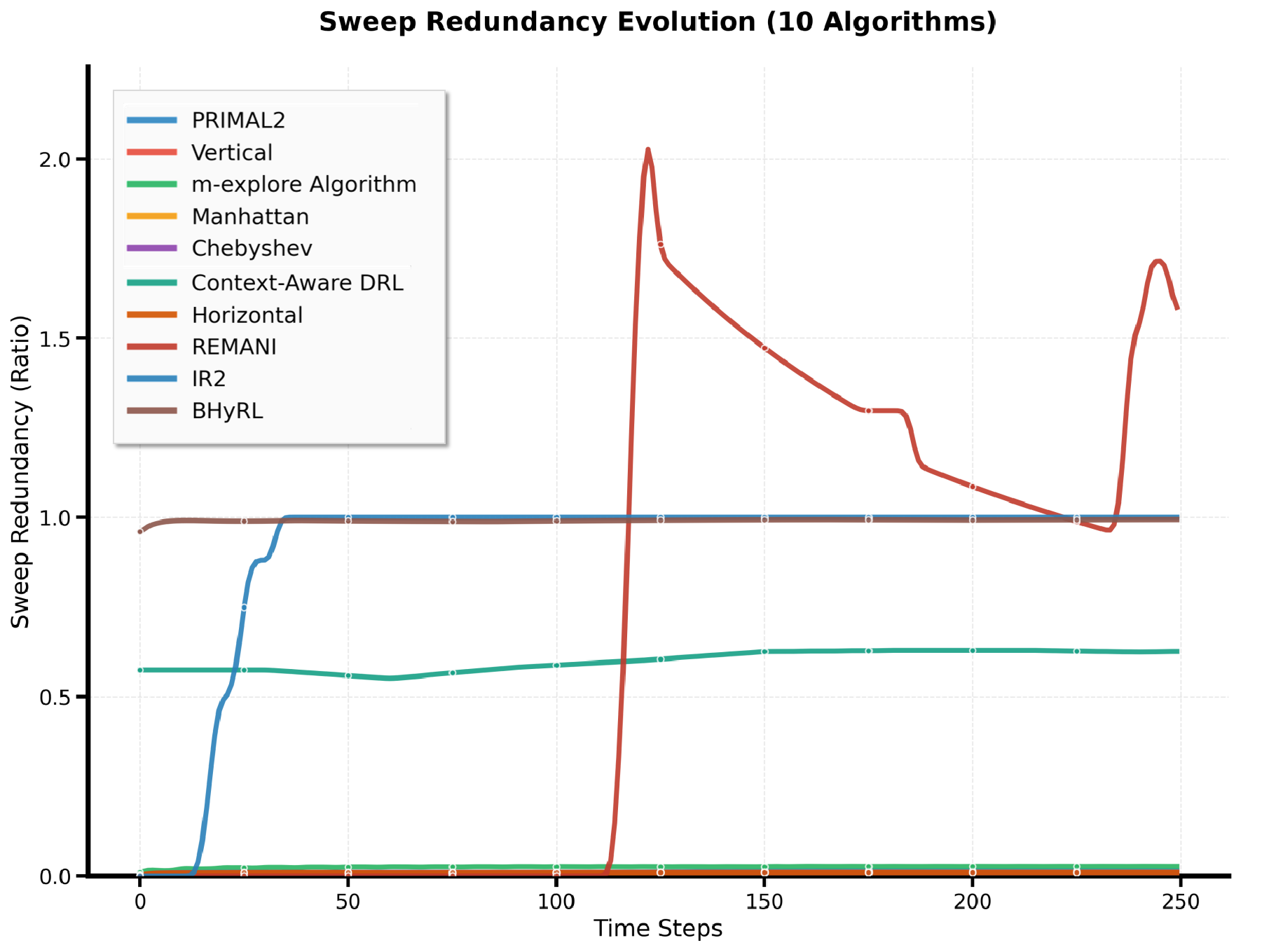}
\caption{Sweep Redundancy Evolution}
\label{fig:sr_evolution}
\end{subfigure}
\caption{Spatial efficiency metrics evolution. Motion efficiency tracks distance per target achieved, while sweep redundancy measures unnecessary area revisitation patterns.}
\label{fig:efficiency_evolution}
\end{figure*}

The efficiency analysis in Figure \ref{fig:efficiency_evolution} demonstrates that Different methods exhibit varying motion efficiency characteristics based on their planning strategies and coordination mechanisms.

\subsection{E.5 Safety and Collision Analysis}

\begin{figure*}[htbp!]
\centering
\begin{subfigure}[b]{0.48\textwidth}
\includegraphics[width=\textwidth]{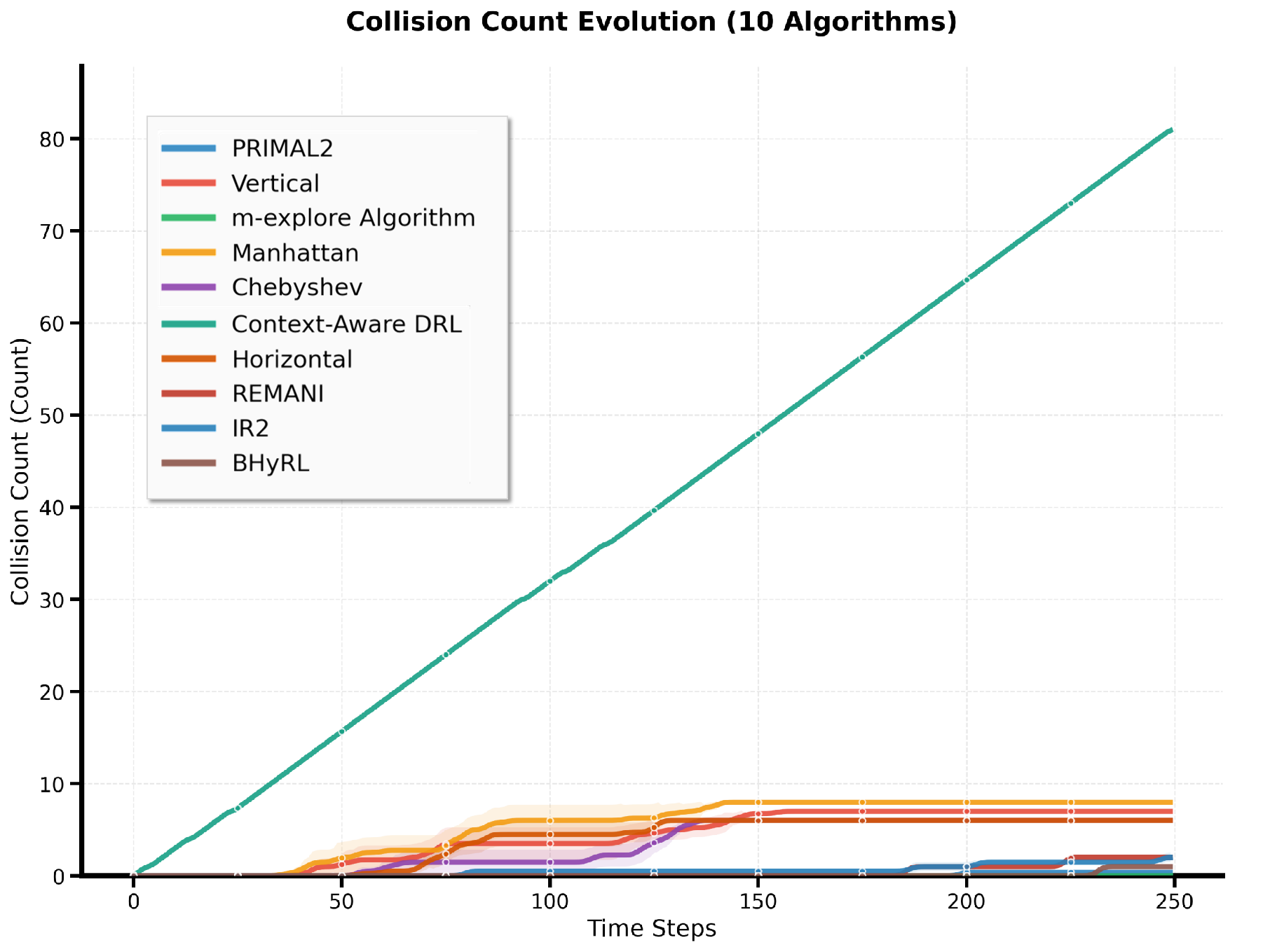}
\caption{Collision Count Evolution}
\label{fig:collision_evolution}
\end{subfigure}
\hfill
\begin{subfigure}[b]{0.48\textwidth}
\includegraphics[width=\textwidth]{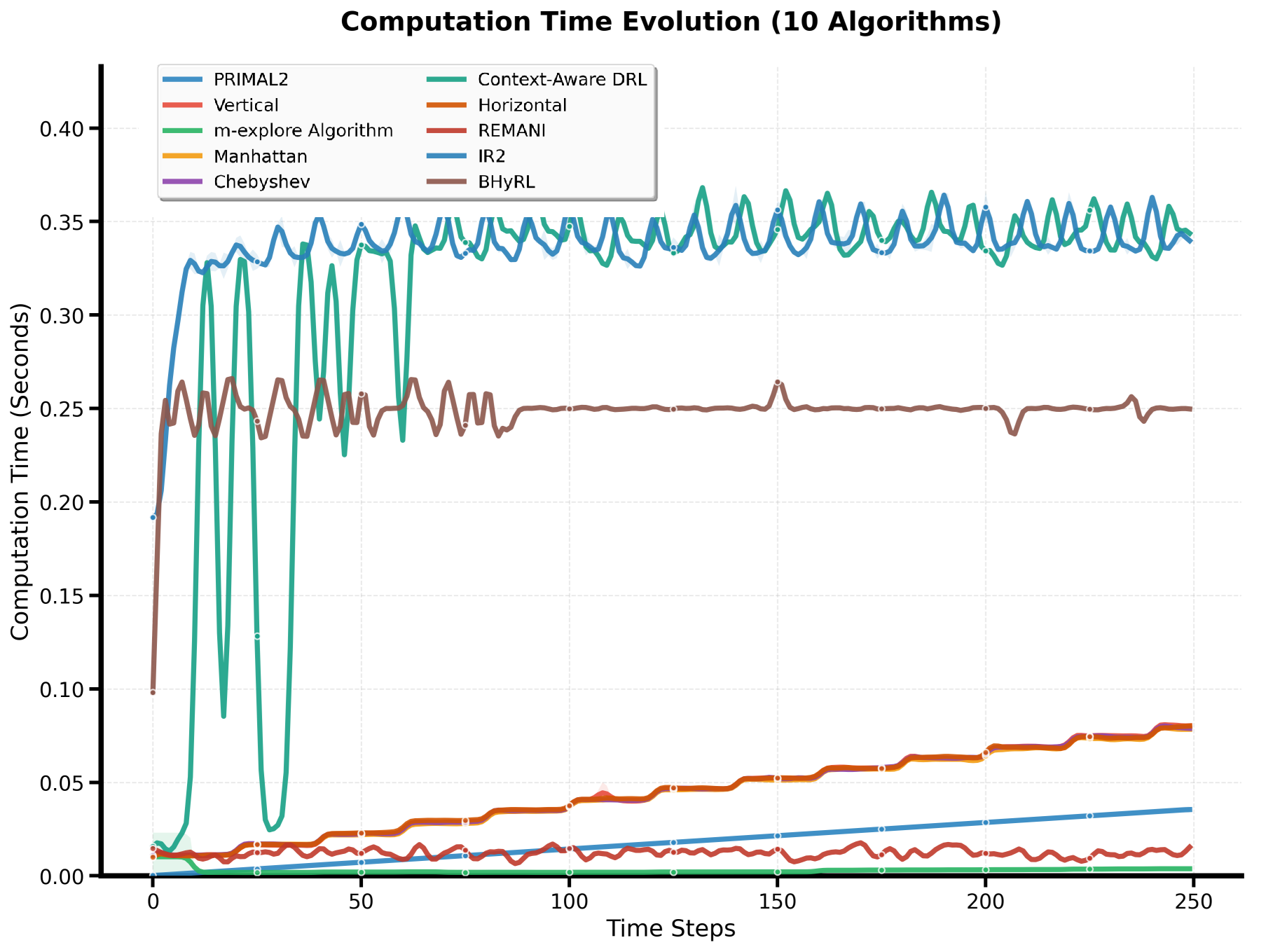}
\caption{Computation Time Evolution}
\label{fig:ct_evolution}
\end{subfigure}
\caption{Safety and computational performance metrics. Collision evolution shows cumulative safety violations while computation time tracks algorithmic efficiency over episode duration.}
\label{fig:safety_computation}
\end{figure*}

Safety analysis \cite{qu2025dpgp} in Figure \ref{fig:safety_computation}(a) reveals that Context-Aware DRL accumulates collision counts linearly, reaching 80+ collisions by episode end, indicating fundamental navigation control issues. In contrast, PRIMAL2 and most heuristic methods maintain near-zero collision counts throughout episodes. Computational analysis shows that Context-Aware DRL and IR2 require 0.34 seconds per decision, while PRIMAL2 achieves 0.04 seconds, demonstrating superior real-time performance.

\subsection{E.6 Task Completion Time Analysis}

\begin{figure*}[htbp!]
\centering
\includegraphics[width=0.8\linewidth]{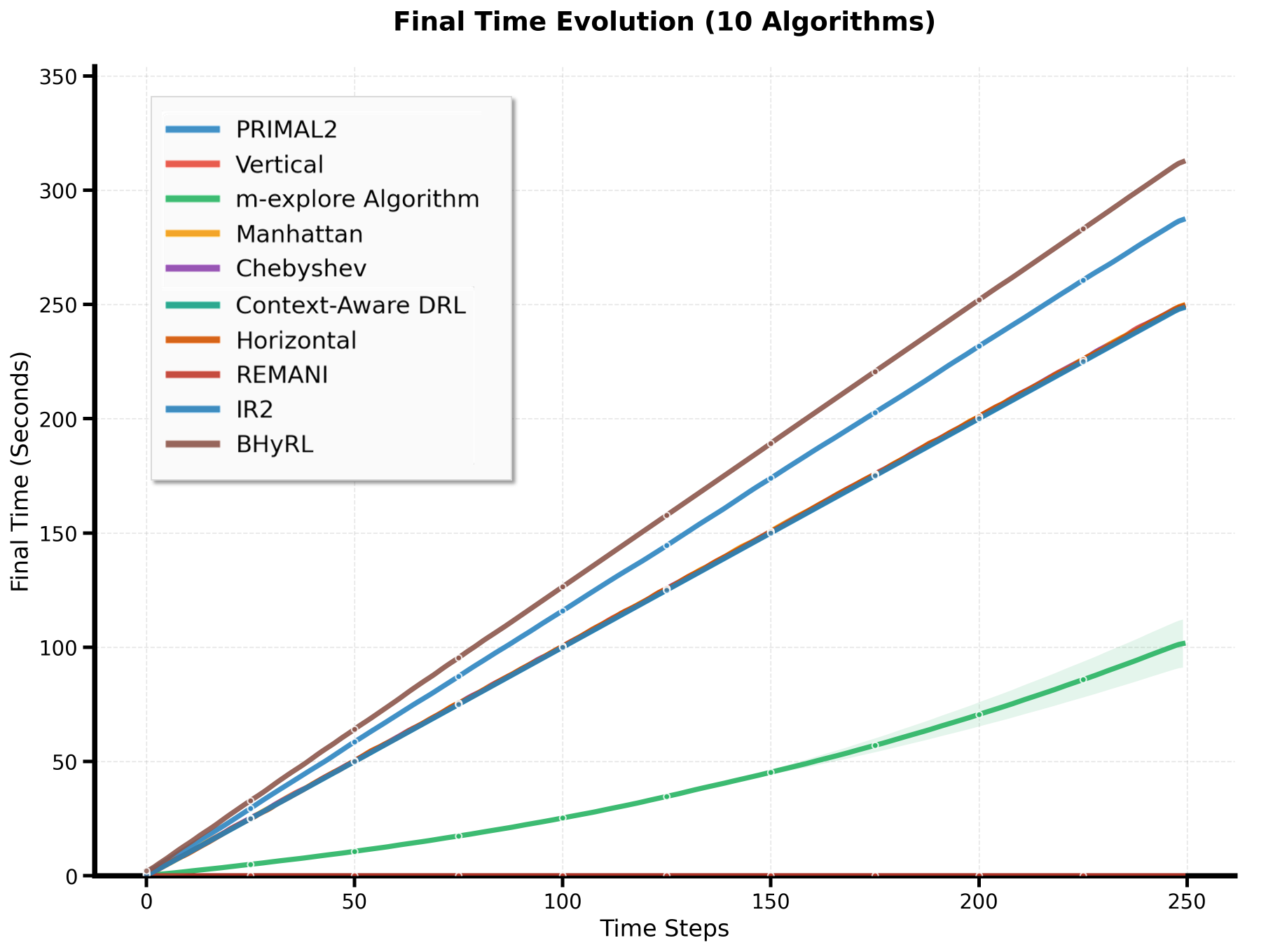}
\caption{Finish time evolution showing cumulative task completion duration. Most algorithms converge to the maximum time limit (250 seconds), while m-explore Algorithm shows variable completion times due to early task completion in simple scenarios.}
\label{fig:ft_evolution}
\end{figure*}

Task completion time analysis in Figure \ref{fig:ft_evolution} shows that most algorithms utilize the full episode duration (250 seconds), indicating that the time constraint is appropriately challenging. m-explore Algorithm demonstrates the most variable completion times, with some episodes finishing as early as 50 seconds in sparse environments, though with minimal task completion success.

\section{F. Robustness and Ablation Studies}

\subsection{F.1 Cross-Scene Generalization}

Cross-scene generalization analysis reveals varying degrees of transferability across different algorithmic approaches. Learning-based methods demonstrate superior adaptation \cite{cao2024reliable} capabilities when evaluated on unseen scene configurations, while heuristic approaches show more predictable but limited performance transfer.

\subsection{F.2 Object Density Variation}

Performance sensitivity analysis across varying object \cite{liu2025handle} densities indicates that most methods experience degraded task completion rates as environmental complexity increases. However, the rate of degradation varies significantly, with PRIMAL2 and REMANI showing the most robust performance under high-density conditions.

\subsection{F.3 Algorithm Performance Correlation Analysis}

\begin{figure*}[htbp!]
\centering
\includegraphics[width=0.95\linewidth]{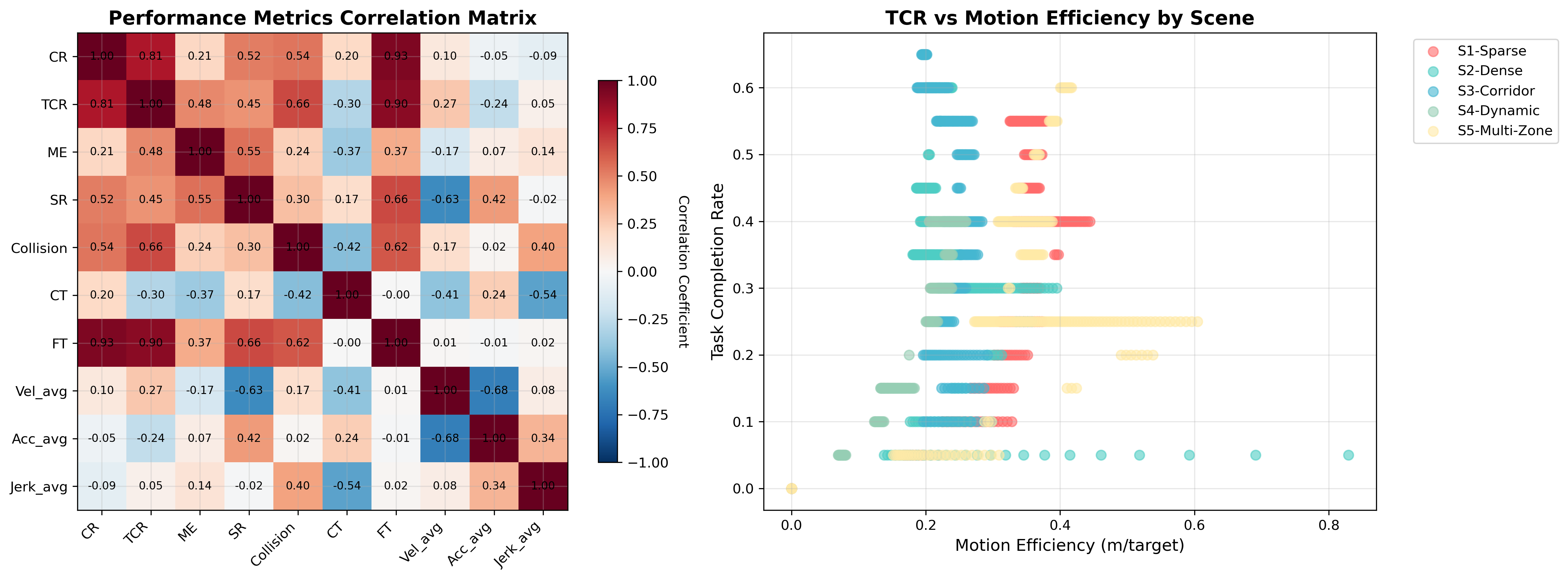}
\caption{Comprehensive correlation matrix between all performance metrics revealing algorithmic relationships and trade-offs. Strong correlations indicate fundamental performance dependencies in dual-mode cleaning tasks.}
\label{fig:correlation_analysis}
\end{figure*}

The correlation analysis in Figure \ref{fig:correlation_analysis} reveals several critical insights about dual-mode cleaning performance relationships:

\textbf{Key Performance Correlations:}
\begin{itemize}
    \item Coverage Rate vs Task Completion Rate (r = 0.81): Strong positive correlation confirms that spatial exploration \cite{bai2024multi} is fundamental for successful dual-mode cleaning
    \item Motion Efficiency vs Sweep Redundancy (r = -0.37): Negative correlation indicates that efficient path planning reduces unnecessary area revisitation
    \item Collision Count vs Task Completion Rate (r = 0.54): Positive correlation suggests that more aggressive exploration strategies may increase collision risk but improve task completion
    \item Finish Time vs Coverage Rate (r = -0.30): Negative correlation shows that algorithms spending more time achieve better spatial coverage
\end{itemize}

\subsection{F.4 Comprehensive Performance Heatmap Analysis}

\begin{figure*}[htbp!]
\centering
\includegraphics[width=0.95\linewidth]{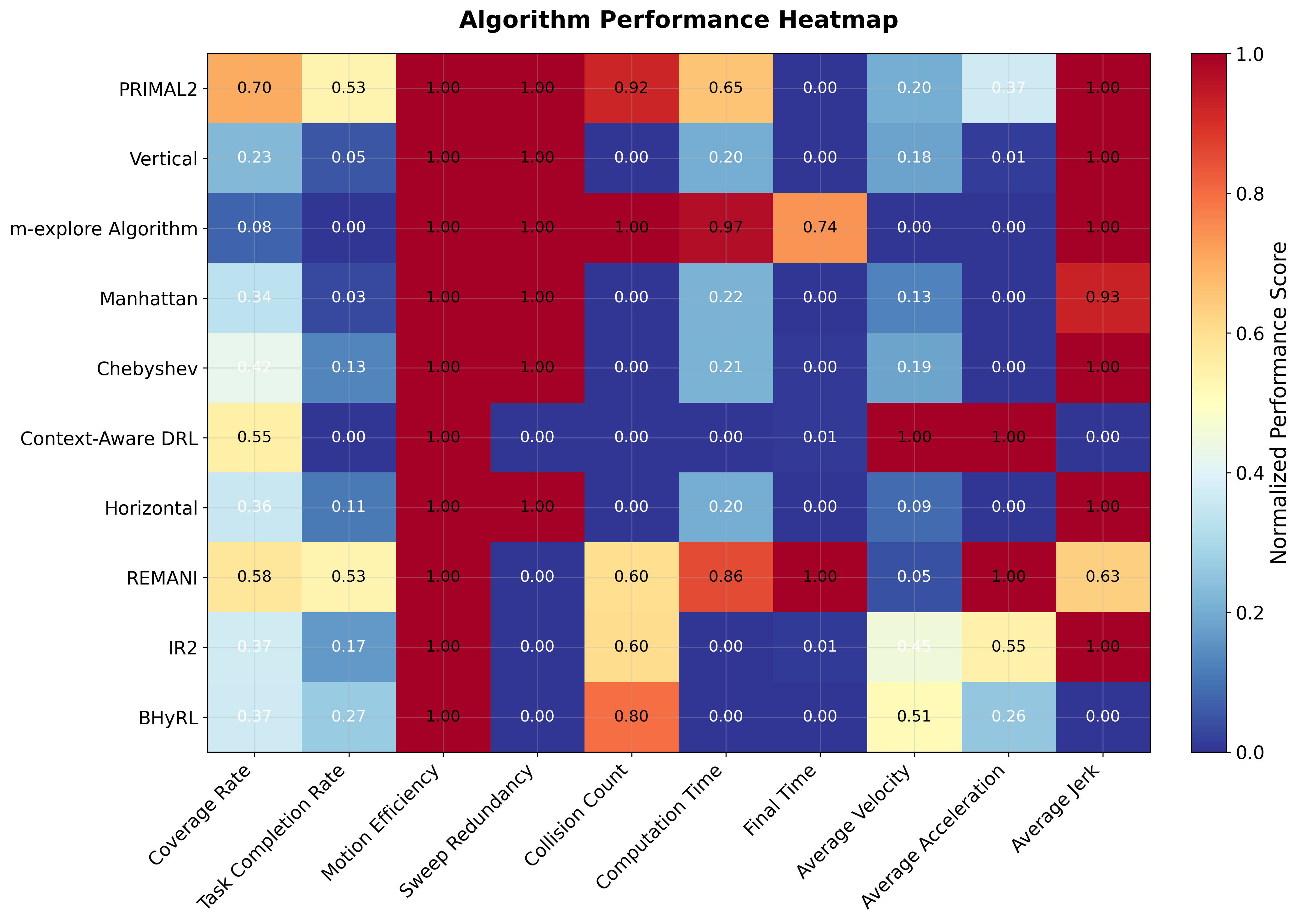}
\caption{Algorithm performance heatmap showing normalized performance scores across all evaluation metrics. Red indicates superior performance while blue shows poor performance, enabling rapid algorithm comparison across multiple dimensions.}
\label{fig:performance_heatmap}
\end{figure*}

The performance heatmap in Figure \ref{fig:performance_heatmap} provides a comprehensive algorithmic comparison across all metrics. PRIMAL2 shows consistently high performance (red coloring) across most metrics including Coverage Rate (0.70) and Task Completion Rate (0.53). Context-Aware DRL exhibits poor performance across most dimensions, while different methods demonstrate varying efficiency characteristics based on their algorithmic approaches and coordination mechanisms.

The heatmap reveals three distinct algorithmic clusters:
1. High-Performance Cluster: PRIMAL2 - balanced excellence across metrics
2. Specialized Efficiency Cluster: REMANI, IR2 - optimized for specific metrics  
3. Safety-Focused Cluster: m-explore Algorithm - excellent collision avoidance but limited task completion
4. Poor Performance Cluster: Context-Aware DRL - consistently low performance across most metrics

\section{G. Real-World Deployment and Future Applications}

\subsection{G.1 Sim-to-Real Transfer Considerations}

CleanUpBench is designed with real-world deployment in mind. Key considerations for sim-to-real transfer include:

\begin{itemize}
    \item \textbf{Domain Randomization}: Isaac Sim supports systematic variation of lighting, textures, and object properties
    \item \textbf{Physics Fidelity}: Accurate contact dynamics and friction modeling for realistic interaction \cite{lai2025natural}
    \item \textbf{Sensor Modeling}: Realistic depth noise and occlusion patterns matching hardware sensors
    \item \textbf{Actuation Limits}: Conservative velocity and acceleration constraints reflecting real robot capabilities
\end{itemize}

\begin{table*}[htbp!]
\centering
\caption{Notation and Symbols Used in CleanUpBench}
\label{tab:notation}
\footnotesize
\renewcommand{\arraystretch}{1.2}
\begin{tabular}{|c|c|l|}
\hline
\textbf{Symbol} & \textbf{Domain/Dimension} & \textbf{Description} \\
\hline
$\mathcal{E}$ & $(\mathcal{S}, \mathcal{A}, \mathcal{T})$ & Cleaning Environment \\
$\mathcal{S}$ & State Space & Environment State Space \\
$\mathcal{A}$ & Action Space & Agent Action Space \\
$\mathcal{T}$ & $\mathcal{S} \times \mathcal{A} \rightarrow \mathcal{S}$ & State Transition Function \\
$s_\tau$ & $\mathcal{S}$ & Environment State at Time Step $\tau$ \\
$a_\tau$ & $\mathcal{A}$ & Agent Action at Time Step $\tau$ \\
$\mathcal{F}$ & $\mathbb{R}^2$ & Robot Footprint in Horizontal Plane \\
$x_\tau$ & $\mathbb{R}^2$ & Robot Position at Time $\tau$ \\
$\theta_\tau$ & $SO(2)$ & Robot Orientation at Time $\tau$ \\
$\mathcal{C}$ & $\{(x_\tau, \theta_\tau)\}_{\tau=1}^{\tau_{\max}}$ & Robot Configuration Space Trajectory \\
\hline
$\mathcal{W}$ & $\mathbb{R}^2$ & Workspace Boundary \\
$A_{\text{total}}$ & $\mathbb{R}^+$ & Total Navigable Floor Area \\
$A_{\text{covered}}$ & $\mathbb{R}^+$ & Cumulative Area Swept by Robot \\
$\mathbb{I}[\cdot]$ & $\{0,1\}$ & Indicator Function \\
$\mathbb{G}$ & Grid Set & Discrete Grid Decomposition of Environment \\
$\delta$ & $\mathbb{R}^+$ & Grid Resolution \\
$\mu_\tau(g)$ & $\{0,1\}$ & Binary Indicator for Grid Cell Intersection \\
$\nu(g)$ & $\mathbb{N}$ & Visit Count for Grid Cell $g$ \\
\hline
$\mathcal{O}_S$ & Object Set & Set of Sweepable Objects \\
$\mathcal{O}_G$ & Object Set & Set of Graspable Objects \\
$\mathcal{O}_{\text{static}}$ & Object Set & Set of Static Obstacles \\
$N_{\text{S-total}}$ & $\mathbb{N}$ & Total Number of Sweepable Objects \\
$N_{\text{G-total}}$ & $\mathbb{N}$ & Total Number of Graspable Objects \\
$N_{\text{S-success}}$ & $\mathbb{N}$ & Number of Successfully Swept Objects \\
$N_{\text{G-success}}$ & $\mathbb{N}$ & Number of Successfully Grasped Objects \\
\hline
$L_{\text{total}}$ & $\mathbb{R}^+$ & Total Trajectory Path Length \\
$\chi_\tau$ & $\{0,1\}$ & Binary Collision Indicator at Time $\tau$ \\
$t_{\text{comp}}(\tau)$ & $\mathbb{R}^+$ & Computation Time for Action at Time $\tau$ \\
$\tau_{\text{init}}$ & $\mathbb{N}$ & Episode Start Time Step \\
$\tau_{\text{final}}$ & $\mathbb{N}$ & Episode End Time Step \\
$\tau_{\max}$ & $\mathbb{N}$ & Maximum Time Steps per Episode \\
$\Delta t$ & $\mathbb{R}^+$ & Discrete Time Interval \\
\hline
$\alpha, \beta$ & $\mathbb{R}^+$ & Weight Parameters for TCR ($\alpha + \beta = 1$) \\
$T_{\text{max}}$ & $\mathbb{R}^+$ & Maximum Time Budget for Episode \\
$L$, $\rho$, $P$ & Parameters & Layout Type, Obstacle Density, Target Pattern \\
$(S, O, T_s, T_g)$ & Configuration & Scene, Obstacles, Sweepable/Graspable Targets \\
\hline
\multicolumn{3}{|c|}{\textbf{Evaluation Metrics}} \\
\hline
CR & $[0,1]$ & Coverage Ratio \\
TCR & $[0,1]$ & Task Completion Ratio \\
TCR$_{\text{sweep}}$ & $[0,1]$ & Sweep Task Completion Ratio \\
TCR$_{\text{grasp}}$ & $[0,1]$ & Grasp Task Completion Ratio \\
ME & $\mathbb{R}^+$ & Motion Efficiency (m/target) \\
SR & $[0,1]$ & Sweep Redundancy \\
FT & $\mathbb{R}^+$ & Finish Time (seconds) \\
CT & $\mathbb{R}^+$ & Computation Time (seconds) \\
Vel$_{\text{avg}}$ & $\mathbb{R}^+$ & Average Velocity (m/s) \\
Acc$_{\text{avg}}$ & $\mathbb{R}^+$ & Average Acceleration (m/s²) \\
Jerk$_{\text{avg}}$ & $\mathbb{R}^+$ & Average Jerk (m/s³) \\
Collision & $\mathbb{N}$ & Total Collision Count \\
\hline
\end{tabular}
\end{table*}

\subsection{G.2 Extension to Multi-Agent Scenarios}

Future extensions will incorporate multi-agent \cite{chen2025relative} coordination scenarios, enabling evaluation of collaborative cleaning strategies, task allocation mechanisms, and communication-based coordination protocols.

\subsection{G.3 Integration with Foundation Models}

Future extensions will incorporate large language models for natural language instruction following and vision-language models for open-vocabulary object recognition, enabling more flexible and human-interpretable cleaning behaviors.

\subsection{G.4 Limitations}

While CleanUpBench provides a unified, extensible benchmark for embodied cleaning agents, several aspects remain simplified relative to real-world deployment settings:

\textbf{Environmental Scope}: Current scenes represent idealized indoor environments with limited diversity in textures, lighting conditions, and dynamic elements compared to real households.

\textbf{Physical Interaction}: Binary affordance labels and deterministic interaction outcomes omit complexities such as deformable objects, partial cleaning, and tool degradation.

\textbf{Sensor Limitations}: Idealized RGB-D sensing without realistic noise, motion blur, or environmental interference that characterizes real-world perception systems.

\subsection{G.5 Future Work}

\textbf{Enhanced Realism}: Integration of soft-body simulation, fluid dynamics, and probabilistic contact models to better represent real-world cleaning scenarios.

\textbf{Language Integration}: Natural language instruction following and semantic \cite{yang2022overcoming} grounding for human-robot interaction evaluation.

\textbf{Long-Horizon Planning}: Support for hierarchical task decomposition \cite{liao2025following} and multi-session \cite{ma2024mm,deng2025mne,deng2025mcn} cleaning scenarios spanning extended time horizons \cite{li2025ll}.

\section{H. Code Availability and Reproducibility}

\subsection{H.1 Open Source Release}

Upon paper acceptance, the complete CleanUpBench codebase will be released under the MIT License, including:

\begin{itemize}
    \item Isaac Sim scene configurations and object assets
    \item Baseline method implementations with hyperparameter settings
    \item Evaluation framework and metric computation scripts
    \item Data visualization and analysis tools
    \item Comprehensive documentation and tutorials
\end{itemize}

\textbf{Repository Structure:}
\begin{verbatim}
CleanUpBench/
|-- environments/# Scene configurations
|-- assets/      # Models and textures  
|-- agents/      # Baseline Deployments
|-- evaluation/  # Metrics and analysis
|-- docs/        # Documentation
+-- scripts/     # Utility scripts
\end{verbatim}

\subsection{H.2 Reproducibility Checklist}

All experimental results are fully reproducible using:
\begin{itemize}
    \item Fixed random seeds for consistent initialization
    \item Detailed hyperparameter specifications for all methods
    \item Standardized evaluation protocols across all experiments
    \item Version-controlled environment configurations
    \item Docker containers for dependency management
\end{itemize}

\subsection{H.3 Community Contributions}

We welcome community contributions to expand CleanUpBench's capabilities through additional baseline methods, scene configurations, evaluation metrics, and real-world validation datasets.

\section{I. Notation and Symbol Definitions}

All mathematical notation and symbols used throughout CleanUpBench are systematically defined in Tab. \ref{tab:notation} for reference and clarity.

\bibliography{aaai2026}


\end{document}